\PassOptionsToPackage{table}{xcolor}
\documentclass{article}

\usepackage{microtype}
\usepackage{graphicx}
\usepackage{subfigure}
\usepackage{booktabs} %

\usepackage{hyperref}

\usepackage[accepted]{icml2024}

\usepackage{amsmath}
\usepackage{amssymb}
\usepackage{mathtools}
\usepackage{amsthm}

\usepackage[capitalize,noabbrev]{cleveref}

\theoremstyle{plain}

\theoremstyle{definition}

\theoremstyle{remark}

\usepackage[textsize=tiny]{todonotes}

\usepackage{booktabs}
\usepackage{multicol}
\usepackage{multirow}
\usepackage{color}
\usepackage[tableposition=top]{caption}
\usepackage{subcaption}
\usepackage{hhline}
\usepackage{pifont}
\usepackage{threeparttable}
\usepackage{makecell}
\usepackage{algorithm} 
\usepackage{listings}
\usepackage{amsfonts,amssymb}
\usepackage{amsmath}
\usepackage{graphicx}
\usepackage{lipsum}
\usepackage{arydshln}
\usepackage[accsupp]{axessibility}
\usepackage{enumitem}
\usepackage{tcolorbox}

\newcommand{\benchname}{MM-Vet}
\def\ie{\emph{i.e.}}
\def\eg{\emph{e.g.}}
\def\etc{\emph{etc}}

\newcommand{\myPara}[1]{\vspace{.05in}\noindent\textbf{#1.}}
\newlength\savedwidth
\newcommand\whline{\noalign{\global\savedwidth\arrayrulewidth\global\arrayrulewidth 0.8pt}\hline\noalign{\global\arrayrulewidth\savedwidth}}
\usepackage{verbatim}

\usepackage{colortbl}
\newcommand{\greentext}[1]{\textcolor[rgb]{0, 0.5, 0}{#1}}
\newcommand{\orangetext}[1]{\textcolor[RGB]{255, 69, 0}{#1}}
\newcommand{\bluetext}[1]{\textcolor[RGB]{0, 0, 255}{#1}}
\newcommand{\purpletext}[1]{\textcolor[RGB]{148, 0, 211}{#1}}

\newcommand{\firstc}{{\cellcolor[RGB]{ 171, 235, 198 }}}
\newcommand{\secondc}{{\cellcolor[RGB]{ 253, 235, 208}}}
\newcommand{\thirdc}{{\cellcolor[RGB]{  214, 234, 248  }}}

\usepackage[T1]{fontenc}

\usepackage{placeins}

\usepackage{float}
\usepackage{stfloats}

\usepackage[figuresright]{rotating}

\icmltitlerunning{MM-Vet: Evaluating Large Multimodal Models for Integrated Capabilities}

\begin{document}

\twocolumn[
\icmltitle{MM-Vet: Evaluating Large Multimodal Models for Integrated Capabilities}

\icmlsetsymbol{equal}{*}

\begin{icmlauthorlist}
\icmlauthor{Weihao Yu}{equal,nus}
\icmlauthor{Zhengyuan Yang}{equal,ms}
\icmlauthor{Linjie Li}{ms}
\icmlauthor{Jianfeng Wang}{ms}
\icmlauthor{Kevin Lin}{ms}
\\
\icmlauthor{Zicheng Liu}{ms}
\icmlauthor{Xinchao Wang}{nus}
\icmlauthor{Lijuan Wang}{ms}
\end{icmlauthorlist}

\icmlaffiliation{nus}{National University of Singapore, Singapore}
\icmlaffiliation{ms}{Microsoft Azure AI, USA}

\icmlcorrespondingauthor{Xinchao Wang}{xinchao@nus.edu.sg}
\icmlcorrespondingauthor{Lijuan Wang}{lijuanw@microsoft.com}

\icmlkeywords{Machine Learning, ICML}

\vskip 0.3in
]

\printAffiliationsAndNotice{\icmlEqualContribution} %

\begin{abstract}
We propose \benchname{}\footnote{Short for ``Multimodal Veterinarian.''}, an evaluation benchmark that examines large multimodal models (LMMs) on complicated multimodal tasks. Recent LMMs have shown various intriguing abilities, such as solving math problems written on the blackboard, reasoning about events and celebrities in news images, and explaining visual jokes.
Rapid model advancements pose challenges to evaluation benchmark development. Problems include: (1) How to systematically structure and evaluate the complicated multimodal tasks; (2) How to design evaluation metrics that work well across question and answer types; and (3) How to give model insights beyond a simple performance ranking. To this end, we present \benchname{}, designed based on the insight that the intriguing ability to solve complicated tasks often stems from a generalist model being able to integrate different core vision-language (VL) capabilities. \benchname{}~defines $6$ core VL capabilities and examines the $16$ integrations of interest derived from their combinations. 
For evaluation metrics, we propose an LLM-based evaluator for open-ended outputs. The evaluator enables the evaluation across different question types and answer styles, resulting in a unified scoring metric. We evaluate representative LMMs on \benchname{}, providing insights into the capabilities of different LMM system paradigms and model designs. Code and data are available at \url{https://github.com/yuweihao/MM-Vet}, and the online evaluator at \url{https://huggingface.co/spaces/whyu/MM-Vet_Evaluator}.
\end{abstract}

\section{Introduction}
\begin{figure}[t]
  \centering
    \includegraphics[width=1.05\linewidth]{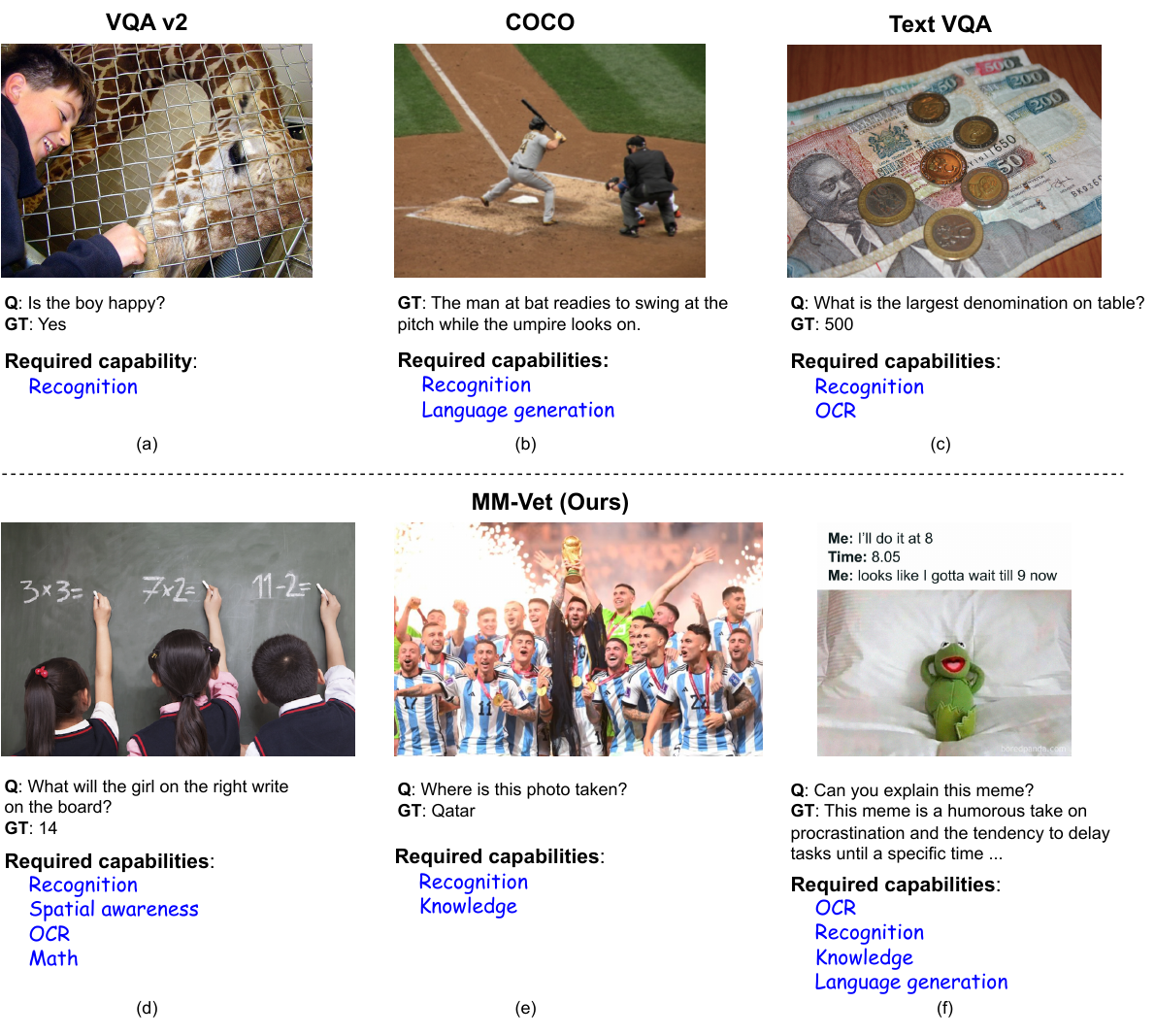}
   \caption{
   The benchmarks differ in their required capabilities. While standard VL benchmarks~\cite{chen2015microsoft,VQA_15,textvqa} typically require only one or two capabilities, \benchname{} focuses on the integration of multiple core VL capabilities. These include recognition, OCR, knowledge, language generation, spatial awareness, and math. 
   }
   \label{fig:first_figure}
\end{figure} 

The breakthroughs in large language models (LLMs)~\cite{gpt3,openai2023gpt4,chowdhery2022palm,anil2023palm,touvron2023llama,hoffmann2022training} bring generalist AI models that can solve a wide range of complicated natural language tasks, many approaching the human-expert-level performance~\cite{openai2023gpt4,bubeck2023sparks,yang2023dawn}. 
Large multimodal models (LMMs) aim to achieve even stronger general intelligence via extending LLMs with multimodal inputs. Since more than 80\% of our human being's perception, learning, cognition, and activities are mediated through vision \cite{vision80}, it is natural to start the exploration by equipping LLMs with ``eyes.''
One main thread of LMM works, represented by Frozen~\cite{tsimpoukelli2021multimodal}, Flamingo~\cite{alayrac2022flamingo}, PaLM-E~\cite{driess2023palme}, GPT-4V~\cite{openai2023gpt4,gpt4v}, extend LLMs with the visual understanding capability via end-to-end tuning. There also exists the exploration~\cite{yang2022empirical,zeng2022socratic,yang2023mm,shen2023hugginggpt,gao2023assistgpt} on the modular combination of LLMs and image-to-text vision-language models.
Recently, thanks to the open-source of powerful LLMs like LLaMA~\cite{touvron2023llama}, more open-sourced LMMs are built, including OpenFlamingo~\cite{anas_awadalla_2023_7733589}, LLaVA~\cite{llava}, MiniGPT-4~\cite{zhu2023minigpt}, Otter \cite{li2023otter}, InstructBLIP~\cite{dai2023instructblip}, and many more~\cite{gong2023multimodalgpt,liu2023visual,ye2023mplug}. These studies showcase the intriguing ability to solve various complicated multimodal tasks, such as open-world recognition, multimodal knowledge and commonsense, scene text understanding, and so on.

Despite the promising qualitative results on LMM's capabilities, it remains unclear how to systematically evaluate those showcased complicated multimodal tasks, and what are the relationships among evaluated tasks, which is the first step in developing a quantitative evaluation benchmark.
As shown in Figure \ref{fig:first_figure}, existing VL benchmarks~\cite{VQA_15,chen2015microsoft,textvqa} focus on straightforward VL tasks that test specific one or two capabilities, such as recognition, language generation, or OCR, but fall short in benchmarking more complicated tasks. In contrast, we examine the integration of multiple core VL capabilities for more complicated tasks. This is based on the insight that the intriguing ability to solve complicated multimodal tasks can be achieved by a generalist model mastering and integrating these core capabilities.
Following this insight, we propose a new benchmark for evaluating LMMs, namely \benchname{}. \benchname{} defines six core VL capabilities, including recognition, OCR, knowledge, language generation, spatial awareness, and math, which integrate to solve various complicated multimodal tasks. \benchname{} contains 16 tasks for quantitative evaluation.
For example, in Figure \ref{fig:first_figure}(d), answering the question ``\textit{What will the girl on the right write on the board?}'' in \benchname{} requires recognizing the genders of the three kids, locating queried girl spatially, recognizing the scene text written by the girl, and finally calculating the result. %

Other than the evaluation category topology, finding effective evaluation metric is another challenge in benchmark development, given the diverse answer styles and question types. Specifically: \textbf{(1)} The desired outputs in different multimodal tasks have diverse formats, \eg, Figure \ref{fig:first_figure}(d)'s math problem can be answered by a single word, while outputs for the essay writing question are hundred-words long; \textbf{(2)} The core aspect to evaluate in different tasks varies, \eg, text generation focuses more on the text quality, recognition can be considered correct with the key concept recognized. Most integrated tasks would require comprehensive evaluations from multiple dimensions.
Inspired by recent NLP studies~\cite{chiang2023can,liu2023gpteval,fu2023gptscore} that use LLMs for model evaluation, we propose an LLM-based evaluator as the evaluation metric for open-ended model outputs. As shown in Table~\ref{tab:prompt}, we prompt GPT-4~\cite{openai2023gpt4} with few-shot evaluation prompts to obtain an evaluation score ranging from $0$ to $1$, conditioned on the question, prediction, and GT annotation. Instead of manually defining the possible answer styles and question types, we include different sample types as few-shot examples and let LLMs infer the scoring criteria automatically. Such metric design eases the future extension to more question types, such as box localization~\cite{chen2021pix2seq,yang2022unitab,wang2023visionllm}.

\benchname{}'s evaluation category and metric designs allow users to obtain per-capability insights for different LMMs. Such model analyses can be more informative than a single overall ranking, which highly depends on the dataset sample composition and might be biased. We evaluate two sets of multimodal systems, \ie, the end-to-end tuned LMMs including OpenFlamingo~\cite{anas_awadalla_2023_7733589}, LLaVA \cite{llava}, MiniGPT-4~\cite{zhu2023minigpt},  Otter~\cite{li2023otter}, InstructBLIP~\cite{dai2023instructblip}, \etc, and the LLM-tool-using systems~\cite{yang2023mm,shen2023hugginggpt,gao2023assistgpt, transformers_agent} such as MM-ReAct~\cite{yang2023mm} and Transformers Agent~\cite{transformers_agent}. Despite not knowing model details, we also evaluate industry solutions such as GPT-4V~\cite{gpt4v} and Bard~\cite{bard}, which are separately tagged to avoid unfair direct comparisons. We first discuss the capability analyses of these two system paradigms and their representative models. We then dive deeper into the open-sourced LMMs and examine how the training data, vision encoder, and LLM selection influence the performance on different capabilities.

\vspace{1pt}
Our contributions are summarized as follows.
\begin{itemize}
\item We propose \benchname{} to evaluate LMMs' ability on complicated multimodal tasks. \benchname{} considers $16$ emergent tasks, integrated from $6$ defined core VL capabilities.
\item We propose an LLM-based evaluator for open-ended outputs from LMMs, which unifies the evaluation across different answer styles and question types. The evaluation metrics ensure the thorough evaluation of both the factual correctness and text quality of the responses.
\item We benchmark representative LMMs on \benchname{}, revealing the relative strengths and weaknesses of different system paradigms and models, as summarized in Section~\ref{sec:takeaway}.
\end{itemize}

\section{Related work}
\noindent\textbf{Multimodal models.}
Vision-language models~\cite{chen2015microsoft,vqav2,lu2019vilbert,chen2019uniter,li2020oscar,kim2021vilt,wang2021simvlm,wang2022git,yang2022unitab,gan2022vision} approach multimodal intelligence of jointly understanding and generating vision and language signals. Inspired by the impressive quality and genericity in recent large language models (LLMs)~\cite{brown2020language,openai2023gpt4,chowdhery2022palm,touvron2023llama}, researchers explore large multimodal models (LMMs) that seamlessly integrate different vision-language capabilities to solve complicated multimodal tasks. In approaching such multimodal generalist systems, one direction is to extend LLMs with the multi-sensory ability, such as pioneer works Frozen~\cite{tsimpoukelli2021multimodal}, Flamingo~\cite{alayrac2022flamingo}, PaLM-E~\cite{driess2023palme} and GPT-4V~\cite{openai2023gpt4, gpt4v}. Recent open-sourced LLMs~\cite{zhang2022opt,touvron2023llama,peng2023instruction} also facilitate various research studies including OpenFlamingo~\cite{anas_awadalla_2023_7733589}, LLaVA~\cite{llava}, MiniGPT-4~\cite{zhu2023minigpt}, Otter \cite{li2023otter}, InstructBLIP~\cite{dai2023instructblip}, and so on~\cite{gong2023multimodalgpt,liu2023visual,ye2023mplug}. On the other hand, multimodal agents~\cite{yang2023mm,shen2023hugginggpt,transformers_agent,gao2023assistgpt} explore chaining different vision tools with LLMs~\cite{brown2020language,openai2023gpt4} to achieve integrated vision-language capabilities.

\noindent\textbf{VL benchmarks.}
Classic VL benchmarks focus on specific capabilities of interest, such as visual recognition~\cite{vqav2}, image description~\cite{chen2015microsoft,agrawal2019nocaps}, as well as other benchmarks for specialized capabilities such as scene text understanding~\cite{textvqa,sidorov2020textcaps,yang2021tap}, commonsense reasoning~\cite{zellers2019recognition}, and outside knowledge~\cite{marino2019ok}. The recent development of generalist LMMs posts a strong need for modernized VL benchmarks, which contain complicated multimodal tasks that require integrated VL capabilities. 

Our \benchname{} is most related to the concurrent evaluation studies~\cite{fu2023mme,liu2023mmbench,li2023seedbench,xu2023lvlm, liu2023aligning} such as MME and MMBench, which design comprehensive evaluation samples to facilitate the LMM evaluation. One major difference is that \benchname{} defines and studies the integrated VL capabilities, allowing the evaluation to provide insights beyond the overall model ranking.

\noindent\textbf{LLM-based evaluation.}
\benchname{} adopts the open-ended LLM-based evaluator, allowing the evaluation across answer styles and question types without requiring binary or multiple answer choices.
The technique of prompting LLMs for model evaluation is related to the explorations in NLP~\cite{chiang2023can,liu2023gpteval,fu2023gptscore}. We show that the technique extends well to multimodal tasks, and presents a unified prompt to evaluate samples with different answer styles and question types.

\begin{figure}[t]
  \centering
   \centerline{\includegraphics[width=1.06\linewidth]{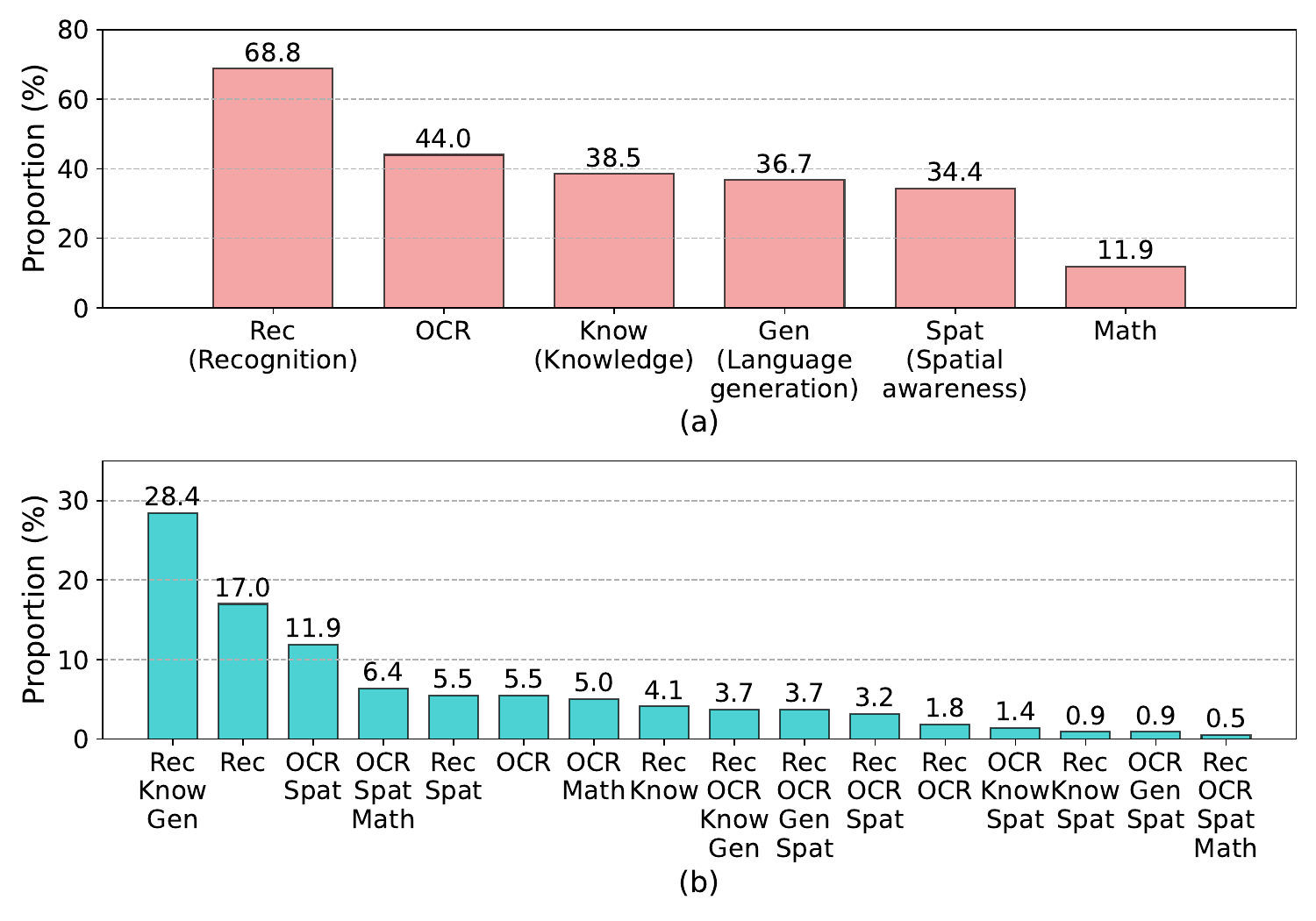}}
   \vspace{-2mm}
   \caption{\benchname{} proportion of capabilities. (a) The proportion of each capability. The sum of the proportion is larger than 100\% because most samples have more than one capability. (b) The proportion of capability integrations. The sum of the proportion is equivalent to 100\%.
   }
   \label{fig:dataset_cap}
   \vspace{-4mm}
\end{figure}

\section{\benchname{}}
\subsection{Data collection}

Our aim is to develop a multimodal benchmark that requires comprehensive capabilities, corresponding to realistic scenarios an AI agent might encounter. Consider, for instance, this scenario: Awakening from slumber, you reach out for your smartphone (\textit{recognition capability}) to check the current time (\textit{OCR capability}). Today, your plan is to visit an unfamiliar grocery store. With the knowledge that it's located opposite a stadium and beside a cinema (\textit{spatial awareness}), you manage to locate it successfully. Remembering your doctor's advice to lose weight, you avoid high-calorie items and instead pick up milk, vegetables, and fruits (\textit{knowledge capability}). In the dairy section, you saw two options of pure milk: one liter at \$4 with a 20\% discount, and 1.5 liters at \$7 with a 25\% discount. After some quick arithmetic, you find the former is cheaper (\textit{math capability}) and select the one-liter carton. Later, as you pass by the cinema, you see someone gesturing towards a poster while introducing a new movie (\textit{language generation}).

From the scenarios of interest, we summarize the following six core VL capabilities for evaluation, with corresponding \benchname{} examples shown in Appendix Tables~\ref{tab:samples1}-\ref{tab:samples6}.
\begin{itemize}
    \item \textbf{Recognition (Rec).} Recognition refers to the general visual recognition capability, including recognizing scenes, objects, object attributes (color, material, shape, \etc), counting, and various other high-level visual recognition tasks in computer vision. 
    \item \textbf{Knowledge (Know).} The knowledge category covers various knowledge-related capabilities, including social and visual commonsense knowledge, encyclopedic knowledge, and time-sensitive knowledge like news. This capability necessitates that the model not only possesses such knowledge, but also effectively utilizes it to solve complicated tasks as required.
    \item \textbf{OCR.} Optical character recognition (OCR) refers to the scene text understanding and reasoning capability. The models are tested to read the scene text in images, and reason over the texts to solve various tasks.
    \item \textbf{Spatial awareness (Spat).} Spatial awareness embodies a diverse spectrum of capabilities related to understanding space, including the comprehension of the spatial relationship among object and scene text regions.
    \item \textbf{Language generation (Gen).} Language generation is a vital ability that empowers models to articulate their responses in a clear, engaging, and informative manner. We use questions that demand more extended answers for language generation capacity evaluation.
    \item \textbf{Math.} Math evaluates the model's arithmetic capability in solving math equations or problems in the wild.
\end{itemize}

{In real-world scenarios, various complicated multimodal tasks would require the integrations of different core VL capabilities. For instance, explaining visual jokes as shown in Appendix Table \ref{tab:samples1}(a) requires recognition, knowledge of humor, and language generation; reading documents and solving math problems as shown in Appendix Table \ref{tab:samples2}(a) takes OCR, spatial awareness and math; and answering exam questions given images as shown in Appendix Table \ref{tab:samples5}(b) needs OCR, knowledge, spatial awareness. To solve these complicated tasks, LMMs are expected to seamlessly integrate different VL capabilities. Therefore, it is crucial to establish a benchmark that evaluates the performance of these integrated abilities within LMMs. }

To build the benchmark, we have gathered 187 images from various online sources and ask 205 questions, each of which requires one or more capabilities to answer. 
As shown in Appendix Tables~\ref{tab:samples1}-\ref{tab:samples6}, these questions are varied in type and entail open-ended responses of differing lengths.
The ground truths for 155 questions are carefully annotated by us to ensure high quality, while the remainder of the answers for 50 questions were gathered from the Internet. In addition to the 187 images, ten extra images with high-quality questions are collected from VCR~\cite{zellers2019recognition}, with the questions and answers modified to an open-ended answering format. Another three images are from ChestX-ray14 \cite{wang2017chestx} to obtain corresponding medical expert knowledge. In total, our \benchname{} contains 200 images, and 218 questions (samples), all paired with their respective ground truths. For each question, we have also identified the capacities required to answer them and displayed this information statistically in Figure~\ref{fig:dataset_cap}.

\begin{table}[t]
\caption{Few-shot prompt for evaluating model outputs using GPT-4, where \bluetext{$\mathcal{Q}$} is a sample's question, \greentext{$\mathcal{G}$} is the ground truth and \greentext{$\mathcal{P}$} is the model output for the sample. In the prompt, there are examples with short and long open-ended answers, enabling the evaluation of diverse answer styles.
Taking the prompt filled with \bluetext{$\mathcal{Q}$}, \greentext{$\mathcal{G}$} and \greentext{$\mathcal{P}$}, GPT-4 will generate a soft grading score from 0 to 1.
}
\label{tab:prompt}
\centering
\centerline{\scalebox{1.05}{
\begin{tcolorbox} 
    \centering
    \scriptsize
    \begin{tabular}{p{0.95\columnwidth}}

Compare the ground truth and prediction from AI models, to give a correctness score for the prediction. <AND> in the ground truth means it is totally right only when all elements in the ground truth are present in the prediction, and <OR> means it is totally right when any one element in the ground truth is present in the prediction. The correctness score is 0.0 (totally wrong), 0.1, 0.2, 0.3, 0.4, 0.5, 0.6, 0.7, 0.8, 0.9, or 1.0 (totally right). Just complete the last space of the correctness score. \\
\\
\bluetext{Question} | \greentext{Ground truth} | \orangetext{Prediction} | \purpletext{Correctness} \\
--- | --- | --- | --- \\
\bluetext{What is x in the equation?} | \greentext{-1 <AND> -5} | \orangetext{x = 3} | \purpletext{0.0} \\
\bluetext{What is x in the equation?} | \greentext{-1 <AND> -5} | \orangetext{x = -1} | \purpletext{0.5} \\
\bluetext{What is x in the equation?} | \greentext{-1 <AND> -5} | \orangetext{x = -5} | \purpletext{0.5} \\
\bluetext{What is x in the equation?} | \greentext{-1 <AND> -5} | \orangetext{x = -5 or 5} | \purpletext{0.5} \\
\bluetext{What is x in the equation?} | \greentext{-1 <AND> -5} | \orangetext{x = -1 or x = -5} | \purpletext{1.0} \\
\bluetext{Can you explain this meme?} | \greentext{This meme is poking fun at the fact that the names of the countries Iceland and Greenland are misleading. Despite its name, Iceland is known for its beautiful green landscapes, while Greenland is mostly covered in ice and snow. The meme is saying that the person has trust issues because the names of these countries do not accurately represent their landscapes.} | \orangetext{The meme talks about Iceland and Greenland. It's pointing out that despite their names, Iceland is not very icy and Greenland isn't very green.} | \purpletext{0.4} \\
\bluetext{Can you explain this meme?} | \greentext{This meme is poking fun at the fact that the names of the countries Iceland and Greenland are misleading. Despite its name, Iceland is known for its beautiful green landscapes, while Greenland is mostly covered in ice and snow. The meme is saying that the person has trust issues because the names of these countries do not accurately represent their landscapes.} | \orangetext{The meme is using humor to point out the misleading nature of Iceland's and Greenland's names. Iceland, despite its name, has lush green landscapes while Greenland is mostly covered in ice and snow. The text `This is why I have trust issues' is a playful way to suggest that these contradictions can lead to distrust or confusion. The humor in this meme is derived from the unexpected contrast between the names of the countries and their actual physical characteristics.} | \purpletext{1.0} \\
\bluetext{$\mathcal{Q}$} | \greentext{$\mathcal{G}$} | \greentext{$\mathcal{P}$} |
\end{tabular}
\end{tcolorbox}
}}
\vspace{1mm}
\vspace{-2mm}
\end{table}

\subsection{LLM-based evaluator for open-ended outputs}
Questions and expected responses in \benchname{} are designed to be open-ended to cover the diverse real-world scenarios. This naturally poses a great challenge in terms of model evaluation and metric design.
Drawing inspiration from recent NLP studies~\cite{chiang2023can, zheng2023judging} that utilize LLMs for open-ended evaluations, we leverage GPT-4 to assist evaluation. As shown in Table~\ref{tab:prompt}, we craft a few-shot prompt for model evaluation. The few-shot design allows us to define the scoring metrics via in-context examples and supports easy extension onto new problem sets. 
Specifically, our implemented prompt incorporates five in-context examples with open-ended short answers and two examples with long answers. We cover examples that are fully correct (\ie, 1.0) or incorrect (\ie, 0.0), as well as examples used to define different types of ``partially correct'' responses. The LLM-based evaluator allows any style of model outputs to be evaluated with a unified consistent metric. Furthermore, it also supports easy adaptation to diverse question types and answer styles by simply modifying the evaluation examples.

By inputting the prompt, GPT-4 automatically generates scores for each sample, conditioned on each sample's input question, ground truth, and model output. The score for each sample ranges from 0 to 1. The total scores are computed by 
\begin{equation}
    S = \frac{\sum\limits_{i=1}^Ns_i}{N} \times 100\%,
\end{equation}
where $s_i$ is the score of sample $i$, and $N$ is the sample number. The score regarding each capability or capability integration can be similarly obtained by
\begin{equation}
    S_c = \frac{\sum s_i}{N_c} \times 100\%, \quad i \in C,
\end{equation}
where $C$ is the set of samples requiring a specific capability or capability integration, and $N_c$ is the sample number of the set. 

\section{Evaluation results}

\subsection{Experiment settings}

We utilize \benchname{} to evaluate two types of LMMs, \ie, (1) end-to-end tuned LMMs (OpenFlamingo \cite{alayrac2022flamingo, anas_awadalla_2023_7733589, awadalla2023openflamingo}, BLIP-2 \cite{li2023blip},  LLaVA \cite{llava}, MiniGPT-4 \cite{zhu2023minigpt}, LLaMA-Adapter V2 \cite{gao2023llama}, Otter \cite{li2023otter} and InstructBLIP \cite{dai2023instructblip}); (2) LLM-tool-using methods (MM-ReAct \cite{yang2023mm} and Transformers Agent \cite{transformers_agent}). The summary of these methods is shown in Appendix Table \ref{tab:model_details}. As shown in Table \ref{tab:prompt}, for each sample, we fill the prompt template with its question, ground truth, and output from a specific LMM. By taking the filled prompt into GPT-4, GPT-4 will generate a score from 0 to 1 for the sample. It is found that outputs of GPT-4 still exist variance, although the temperature is set as 0. Therefore, we utilize GPT-4 to evaluate the outputs of LLMs by 5 times. Due to the space limit, we report average scores for capabilities/capability integrations, and average as well as variance for total score.

\subsection{Result analyses}
\label{sec:result}
The main results of different methods are shown in Table \ref{tab:results_cap} regarding each capability, and Table \ref{tab:results_cap_integration} for each capability integration. 

\subsubsection{Regarding each capability}
\vspace{-.05in}
\myPara{Recognition} 
The ``Recognition'' category contains the questions requiring recognition capability to answer. Examples are shown in Appendix Tables \ref{tab:samples1}(a, b), \ref{tab:samples2}(b), \ref{tab:samples3}(a, b), \ref{tab:samples4}(a, b), \ref{tab:samples5}(a, c), and \ref{tab:samples6}(b). The ``Rec'' column in Table~\ref{tab:results_cap} compares the performance on the ``Recognition''.
Among the evaluated models, LLaVA-13B (LLaMA-2) is the best one, obtaining 39.2\%. There may be two reasons. First, LLaVA-13B (LLaMA-2) adopts ViT-L/14 \cite{dosovitskiy2020image} from CLIP \cite{radford2021learning} as a vision model, which is trained by a large amount of data, 400 million image-text pairs; 2) Second, it is surprising that stronger language model can largely boost the recognition performance. LLaVA-13B (LLaMA-2) obtains 8.3\% important over LLaVA-13B (Vicuna-13B). Stronger LLMs may help understand questions better and identify key information from visual inputs.

LLaMA-Adapter v2-7B is another strong model in recognition, achieving 38.5\%. This outstanding ability may be obtained from its various and large amounts of tuning data, LAION-400M \cite{schuhmann2021laion}, COYO-700M \cite{kakaobrain2022coyo-700m}, Multimodal C4 \cite{zhu2023multimodal} and tuning data of LLaVA \cite{llava} \etc~ as shown in Table \ref{tab:model_details}.
Besides, InstructBLIP-8B \cite{dai2023instructblip} attains 32.4\%. As shown in Table \ref{tab:model_details}, the tuning data of InstructBLIP includes 26 publicly available datasets, which contain recognition heavily datasets, like VQA v2 \cite{vqav2} and GQA \cite{hudson2019gqa}. The promising capability of InstructBLIP in recognition may benefit from these datasets.

\begin{table}[t]
  \caption{\benchname{} evaluation results on various LMMs regarding each \textit{core VL capability}. For each column, the highest, second, and third highest figures are highlighted by \setlength{\fboxsep}{0pt}\colorbox[RGB]{ 171, 235, 198 }{green}, \setlength{\fboxsep}{0pt}\colorbox[RGB]{ 253, 235, 208}{orange} and \setlength{\fboxsep}{0pt}\colorbox[RGB]{  214, 234, 248  }{blue} colors. Numbers are presented in \% with a full score of 100\%.
  }
  \vspace{-3mm}
  \setlength{\tabcolsep}{1.6pt}
  \scriptsize
  \centering
    \begin{tabular}{lcccccccc}
    \whline
        Model & Rec & OCR & Know & Gen & Spat & Math & Total  \\ 
        \whline
        Transformers Agent (GPT-4) 
        & 18.2 & 3.9 & 2.2 & 3.2 & 12.4 & 4.0 & 13.4$\pm$0.5 \\
        MiniGPT-4-8B 
        & 27.4 & 15.0 & 12.8 & 13.9 & 20.3 & 7.7 & 22.1$\pm$0.1  \\ 
        BLIP-2-12B 
        &  27.5 & 11.1 & 11.8 & 7.0 & 16.2 & 5.8 & 22.4$\pm$0.2 \\ 
        LLaVA-7B 
        & 28.0 & 17.1 & 16.3 & 18.9 & 21.2 & \thirdc{} 11.5 & 23.8$\pm$0.6 \\ 
        MiniGPT-4-14B 
        & 29.9 & 16.1 & 20.4 & 22.1 & 22.2 & 3.8 & 24.4$\pm$0.4 \\
        Otter-9B 
        & 27.3 & 17.8 & 14.2 & 13.8 & 24.4 & 3.8 & 24.7$\pm$0.3 \\ 
        OpenFlamingo-9B 
        & 28.7 & 16.7 & 16.4 & 13.1 & 21.0 & 7.7 & 24.8$\pm$0.2 \\
        InstructBLIP-14B 
        & 30.8 & 16.0 & 9.8 & 9.0 & 21.1 & 10.5 & 25.6$\pm$0.3 \\
        InstructBLIP-8B 
        & 32.4 & 14.6 & 16.5 & 18.2 & 18.6 & 7.7 & 26.2$\pm$0.2 \\
        LLaVA-13B 
        & 30.9 & 20.1 & 23.5 & 26.4 & 24.3 & 7.7 & 26.4$\pm$0.1 \\
        MM-ReAct-GPT-3.5 
        & 24.2 & \secondc{} 31.5 & 21.5 & 20.7 & \secondc{} 32.3 & \secondc{} 26.2 & 27.9$\pm$0.1 \\ 
        LLaVA-7B (LLaMA-2) 
        & 32.9 & 20.1 & 19.0 & 20.1 & 25.7 & 5.2 & 28.1$\pm$0.4 \\
        LLaMA-Adapter v2-7B 
        & \secondc{} 38.5 & 20.3 & \firstc{} 31.4 & \secondc{} 33.4 & 22.9 & 3.8 & 31.4$\pm$0.1 \\
        LLaVA-13B (V1.3, 336px) 
        & \thirdc{} 38.1 & 22.3 & 25.2 & 25.8 & \thirdc{} 31.3 & 11.2 & \thirdc{} 32.5$\pm$0.1 \\
        LLaVA-13B (LLaMA-2) 
        & \firstc{} 39.2 & \thirdc{} 22.7 & \thirdc{} 26.5 & \thirdc{} 29.3 & 29.6 & 7.7 & \secondc{} 32.9$\pm$0.1 \\
        MM-ReAct-GPT-4 
        & 33.1 & \firstc{} 65.7 & \secondc{} 29.0 & \firstc{} 35.0 & \firstc{} 56.8 & \firstc{} 69.2 & \firstc{} 44.6$\pm$0.2 \\ 
        \whline
    \end{tabular}
  \label{tab:results_cap}
  \vspace{-3mm}
\end{table}

\begin{table*}[t]
  \caption{\benchname{} evaluation results on various LMMs regarding each \textit{capability integration}. Examples of each capability integration are shown in Appendix Tables~\ref{tab:samples1}-\ref{tab:samples6}. For each column, the highest, second, and third highest figures are highlighted by \setlength{\fboxsep}{0pt}\colorbox[RGB]{ 171, 235, 198 }{green}, \setlength{\fboxsep}{0pt}\colorbox[RGB]{ 253, 235, 208}{orange} and \setlength{\fboxsep}{0pt}\colorbox[RGB]{  214, 234, 248  }{blue} colors. Numbers are presented in \% with a full score of 100\%.
  }
  \label{tab:results_cap_integration}
  \vspace{-0.12in}
  \scriptsize
  \setlength{\tabcolsep}{2pt}
  \centering
    \scalebox{1}{
    \begin{tabular}{lcccccccccccccccccc}
    \whline
        \multirow{4}{*}{\makecell[c]{Model}} & \multirow{4}{*}{\makecell[c]{~ \\ Rec \\ Know \\ Gen}} & \multirow{4}{*}{\makecell[c]{~ \\ ~ \\ ~ \\ Rec }} & \multirow{4}{*}{\makecell[c]{~ \\ ~ \\ OCR \\ Spat}} & \multirow{4}{*}{\makecell[c]{~ \\ OCR \\ Spat \\ Math}} & \multirow{4}{*}{\makecell[c]{~ \\ ~ \\ Rec \\ Spat}} & \multirow{4}{*}{\makecell[c]{~ \\ ~ \\ ~ \\ OCR}} & \multirow{4}{*}{\makecell[c]{~ \\ ~ \\ OCR \\ Math}} & \multirow{4}{*}{\makecell[c]{~ \\ ~ \\ Rec \\ Know}} & \multirow{4}{*}{\makecell[c]{Rec \\ OCR \\ Know \\ Gen }}  & \multirow{4}{*}{\makecell[c]{Rec \\ OCR \\ Gen \\ Spat }} & \multirow{4}{*}{\makecell[c]{~ \\ Rec \\ OCR \\ Spat }} & \multirow{4}{*}{\makecell[c]{~ \\ ~ \\ Rec \\ OCR }} & \multirow{4}{*}{\makecell[c]{~ \\ OCR \\ Know \\ Spat }} & \multirow{4}{*}{\makecell[c]{~ \\ Rec \\ Know \\ Spat}}  & \multirow{4}{*}{\makecell[c]{~ \\ OCR \\ Gen \\ Spat }} & \multirow{4}{*}{\makecell[c]{Rec \\ OCR \\ Spat \\ Math }} & \multirow{4}{*}{\makecell[c]{Total}} \\
        ~ \\
        ~ \\
        ~ \\
        \whline
        Transformers Agent (GPT-4) \cite{transformers_agent} & 1.3 & 49.1 & 0.0 & 7.4 & 45.8 & 0.0 & 0.0 & 0.0 & 0.0 & 9.5 & 0.0 & 25.0 & 0.0 & \firstc{} 50.0 & \secondc{} 49.0 & 0.0 & 13.4$\pm$0.5 \\
        MiniGPT-4-8B \cite{zhu2023minigpt} & 14.2 & 47.9 & 9.6 & 14.3 & 50.0 & 20.8 & 0.0 & 14.4 & 8.0 & 21.2 & \firstc{} 42.9 & 50.0 & 0.7 & 0.0 & 0.0 & 0.0 & 22.1$\pm$0.1 \\ 
        BLIP-2-12B \cite{li2023blip} & 7.3 & \thirdc{} 65.1 & 11.5 & 7.1 & 41.7 & 21.2 & 4.5 & \secondc{} 38.9 & 5.2 & 8.5 & 14.3 & 25.0 & 16.7 & \firstc{} 50.0 & 0.0 & 0.0 & 22.4$\pm$0.2 \\ 
        LLaVA-7B \cite{llava} & 17.1 & 46.6 & 13.3 & \thirdc{} 21.4 & 41.7 & 24.8 & 0.0 & 28.9 & 6.2 & 45.2 & 6.6 & 50.0 & 0.0 & 0.0 & 19.0 & 0.0 & 23.8$\pm$0.6 \\
        MiniGPT-4-14B \cite{zhu2023minigpt} & 21.1 & 47.5 & 14.6 & 7.1 & 50.0 & 16.7 & 0.0 & 11.1 & \thirdc{} 18.7 & 38.5 & 18.3 & 32.5 & \thirdc{} 50.0 & 0.0 & 0.0 & 0.0 & 24.4$\pm$0.4 \\
        Otter-9B \cite{li2023otter} & 15.6 & 54.1 & 29.2 & 7.1 & 50.0 & 22.5 & 0.0 & 11.1 & 3.2 & 6.0 & 23.1 & 46.5 & 33.3 & 0.0 & 30.0 & 0.0 & 24.7$\pm$0.3 \\
        OpenFlamingo-9B \cite{awadalla2023openflamingo}
        & 15.5 & 48.6 & 15.4 & 14.3 & \secondc{} 58.3 & 40.5 & 0.0 & \secondc{} 38.9 & 4.5 & 6.0 & \secondc{} 28.6 & 50.0 & 10.0 & 0.0 & 0.0 & 0.0 & 24.8$\pm$0.2 \\ 
        InstructBLIP-14B \cite{dai2023instructblip} & 8.1 & \firstc{} 74.3 & 14.6 & 14.3 & 50.0 & 19.2 & 6.5 & 11.1 & 8.8 & 15.2 & 14.3 & \firstc{} 70.0 & 16.7 & \firstc{} 50.0 & 15.0 & 0.0 & 25.6$\pm$0.3 \\
        InstructBLIP-8B \cite{dai2023instructblip} & 18.0 & \secondc{}{} 69.9 & 15.4 & 14.3 & 33.3 & 20.8 & 0.0 & 23.3 & 7.8 & 35.2 & 15.7 & 25.0 & 0.0 & 0.0 & 0.0 & 0.0 & 26.2$\pm$0.2 \\ 
        LLaVA-13B \cite{llava} & 25.2 & 41.1 & 17.3 & 7.1 & 47.5 & 23.3 & \secondc{} 9.1 & 18.0 & 12.5 & \thirdc{} 53.8 & 14.3 & 50.0 & \thirdc{} 50.0 & 0.0 & 12.0 & 0.0 & 26.4$\pm$0.1 \\
        MM-ReAct-GPT-3.5~\cite{yang2023mm} & 19.1 & 33.1 & \secondc{} 28.8 & \secondc{} 35.7 & 28.3 & \secondc{} 60.0 & \secondc{} 9.1 & 33.3 & 2.5 & 47.8 & 0.0 & 25.0 & \firstc{} 100.0 & 0.0 & \thirdc{} 35.0 & \firstc{} 80.0 & 27.9$\pm$0.1 \\ 
        LLaVA-7B (LLaMA-2) \cite{llava} & 18.8 & 57.0 & \thirdc{} 26.9 & 9.7 & 50.0 & 26.7 & 0.0 & 34.7 & 10.2 & 44.8 & 14.3 & 50.0 & 11.3 & 0.0 & 0.0 & 0.0 & 28.1$\pm$0.4 \\
        LLaMA-Adapter v2-7B \cite{gao2023llama} & \firstc{} 35.3 & 54.1 & 13.5 & 7.1 & 50.0 & \thirdc{} 38.5 & 0.0 & 12.2 & \secondc{} 22.5 & 38.0 & \secondc{} 28.6 & 48.0 & \secondc{} 53.3 & 0.0 & 0.0 & 0.0 & 31.4$\pm$0.1 \\ 
        LLaVA-13B (V1.3, 336px) \cite{llava} & \thirdc{} 25.5 & 59.7 & 25.0 & 14.3 & \firstc{} 66.7 & 25.8 & 8.2 & 27.8 & 11.2 & 49.3 & 14.3 & 50.0 & 33.3 & \firstc{} 50.0 & 2.0 & 0.0 & \thirdc{} 32.5$\pm$0.1 \\
        LLaVA-13B (LLaMA-2) \cite{llava} & \secondc{} 29.8 & 59.5 & 21.2 & 14.3 & \secondc{} 58.3 & 36.2 & 0.0 & 27.8 & 3.5 & \secondc{} 56.8 & \secondc{} 28.6 & 50.0 & 33.3 & 0.0 & 8.0 & 0.0 & \secondc{} 32.9$\pm$0.1 \\
        MM-ReAct-GPT-4~\cite{yang2023mm} & 22.5 & 33.0 & \firstc{} 69.2 & \firstc{} 78.6 & 25.0 & \firstc{} 83.0 & \firstc{} 63.6 & \firstc{} 44.4 & \firstc{} 68.2 & \firstc{} 88.0 & 14.3 & 50.0 & 0.0 & \firstc{} 50.0 & \firstc{} 80.0 & 0.0 & \firstc{} 44.6$\pm$0.2 \\ 
        \whline
    \end{tabular}
    }
\end{table*}

\myPara{OCR} OCR assesses models' capabilities in recognizing scene texts in images and performing various types of reasoning including math, spatial, recognition, \etc.
Examples are shown in Appendix Tables \ref{tab:samples1}(c), \ref{tab:samples2}(a, c, d), \ref{tab:samples3}(b), \ref{tab:samples4}(a, b), \ref{tab:samples5}(a, b), \ref{tab:samples6}(a, b). 
As shown in Table \ref{tab:model_details}'s ``OCR'' column, MMReAct-GPT4~\cite{yang2023mm} performs the best (65.7\%) in OCR capability with the assistance of an external OCR model as a tool. 
Among end-to-end tuned models, LLaVA-13B (LLaMA-2)~\cite{llava} achieves the highest performance (22.7\%). This superior performance may be attributed to LLaVA's adoption of CLIP \cite{radford2021learning} ViT-L/14 \cite{dosovitskiy2020image} as its vision model, and the inclusion of a large volume of image-OCR pairings within the training data~\cite{liu2023hidden}.

\myPara{Knowledge} 
As depicted in Appendix Tables~\ref{tab:samples1}(a), \ref{tab:samples3}(a, b) and \ref{tab:samples5}(b, c), the ``knowledge'' category covers a wide range of knowledge-related questions, ranging from joke understanding to encyclopedia knowledge. 
LLaVA-Adapter v2-7B is the best model in this capability with a score of 31.4\%, as shown in Table \ref{tab:results_cap}. It may be beneficial from its large-scale tuning data including GPT-4-LLM \cite{peng2023instruction}.
MMReAct-GPT-4 \cite{yang2023mm} also achieves a good score (29.0\%) in this capability, because of its strong LLM backbone~\cite{openai2023gpt4}, coupled with external tools like Bing search for knowledge acquisition. 

\myPara{Language generation} ``Language generation'' denotes the proficiency to produce fluent and informative text outputs, as illustrated in Appendix Tables~\ref{tab:samples1}(a), \ref{tab:samples3}(b), \ref{tab:samples4}(a), and \ref{tab:samples6}(a).
The performance within this category is highly correlated with the efficacy of language modeling. As a result, MMReAct-GPT4 \cite{yang2023mm} stands out and its success can be attributed to the GPT-4 on which this system is built.

\myPara{Spatial awareness} ``Spatial awareness'' involves the understanding of the spatial relationship among visual object regions (\eg, Appendix Table~\ref{tab:samples1}(c)) and scene text regions (\eg, Table~\ref{tab:samples4}(a, b)). 
MMReAct-GPT4 \cite{yang2023mm} has a significant lead in this capability (56.8\%), because the adopted tools, such as dense captioning and OCR, provide detailed object and scene text location information in the form of coordinates, which can be understood and processed by GPT-4.

When it comes to end-to-end tuned models, LLaVA-13B (V1.3, 336px) exhibits the best performance of 31.3\%. The tuning data for LLaVA is partly derived from capturing object names and their corresponding coordinates as input. This procedure ensures the generation of data imbued with spatial information, potentially aiding the models in developing and enhancing their spatial awareness capabilities.

\myPara{Math} ``Math'' measures the arithmetic capability on either written equations (\eg, Appendix Table~\ref{tab:samples6}(b)) or problems in the wild (\eg, Appendix Table~\ref{tab:samples2}(d)).
Notably, MMReAct-GPT4 \cite{yang2023mm} consistently outperforms other models. This good performance may be attributed to the adopted PAL math tool (Program-aided Language Models)~\cite{gao2022pal}.

\begin{table*}[h]
  \caption{\benchname{} evaluation of LLaVA, MM-ReAct and GPT-4V regarding each \textit{capability integration}. For each column, the highest and second highest figures are highlighted by \setlength{\fboxsep}{0pt}\colorbox[RGB]{ 171, 235, 198 }{green} and \setlength{\fboxsep}{0pt}\colorbox[RGB]{ 253, 235, 208}{orange} colors. Numbers are presented in \%.
  }
  \label{tab:results_cap_integration_gptv}
  \scriptsize
  \vspace{-0.12in}
  \setlength{\tabcolsep}{2pt}
  \centering  
    \scalebox{1}{
    \begin{tabular}{lcccccccccccccccccc}
    \hline
        \multirow{4}{*}{\makecell[c]{Model}} & \multirow{4}{*}{\makecell[c]{~ \\ Rec \\ Know \\ Gen}} & \multirow{4}{*}{\makecell[c]{~ \\ ~ \\ ~ \\ Rec }} & \multirow{4}{*}{\makecell[c]{~ \\ ~ \\ OCR \\ Spat}} & \multirow{4}{*}{\makecell[c]{~ \\ OCR \\ Spat \\ Math}} & \multirow{4}{*}{\makecell[c]{~ \\ ~ \\ Rec \\ Spat}} & \multirow{4}{*}{\makecell[c]{~ \\ ~ \\ ~ \\ OCR}} & \multirow{4}{*}{\makecell[c]{~ \\ ~ \\ OCR \\ Math}} & \multirow{4}{*}{\makecell[c]{~ \\ ~ \\ Rec \\ Know}} & \multirow{4}{*}{\makecell[c]{Rec \\ OCR \\ Know \\ Gen }}  & \multirow{4}{*}{\makecell[c]{Rec \\ OCR \\ Gen \\ Spat }} & \multirow{4}{*}{\makecell[c]{~ \\ Rec \\ OCR \\ Spat }} & \multirow{4}{*}{\makecell[c]{~ \\ ~ \\ Rec \\ OCR }} & \multirow{4}{*}{\makecell[c]{~ \\ OCR \\ Know \\ Spat }} & \multirow{4}{*}{\makecell[c]{~ \\ Rec \\ Know \\ Spat}}  & \multirow{4}{*}{\makecell[c]{~ \\ OCR \\ Gen \\ Spat }} & \multirow{4}{*}{\makecell[c]{Rec \\ OCR \\ Spat \\ Math }} & \multirow{4}{*}{\makecell[c]{Total}} \\
        ~ \\
        ~ \\
        ~ \\
        \whline
        LLaVA-13B (LLaMA-2) \cite{llava} & 29.8 & 59.5 & 21.2 & 14.3 & 58.3 & 36.2 & 0.0 & 27.8 & 3.5 &  56.8 & 28.6 & 50.0 & 33.3 & 0.0 & 8.0 & 0.0 & 32.9$\pm$0.1 \\
        MM-ReAct-GPT-4~\cite{yang2023mm} & 22.5 & 33.0 & \secondc{} 69.2 & \secondc{} 78.6 & 25.0 & \firstc{} 83.0 & \firstc{} 63.6 & \firstc{} 44.4 & 68.2 & \firstc{} 88.0 & 14.3 & 50.0 & 0.0 & \secondc{} 50.0 & 80.0 & 0.0 & 44.6$\pm$0.2 \\ 
        GPT-4V~\cite{gpt4v} & \firstc{} 55.5 & \firstc{} 89.2 & 68.6 & 73.9 & \firstc{} 83.3 & 77.5 & 44.5 & \secondc{} 38.9 & 78.2 & 76.5 & \firstc{} 42.9 & \firstc{} 100.0 & \firstc{} 66.7 & \secondc{} 50.0 & \secondc{} 89.0 & 0.0 & \firstc{} 67.7$\pm$0.3 \\ 
        GPT-4V-Turbo-detail:low~\cite{gpt4v} & \secondc{} 52.9 & 75.7 & 58.4 & 50.0 & \secondc{} 75.0 & 69.2 & 45.5 & \secondc{} 38.9 & \firstc{} 84.8 & \secondc{} 85.8 & 14.3 & 75.0 & \firstc{} 66.7 & \secondc{} 50.0 & 87.0 & 0.0 & 60.2$\pm$0.3 \\ 
        GPT-4V-Turbo-detail:high~\cite{gpt4v} & 50.2 & \secondc{} 77.7 & \firstc{} 82.5 & \firstc{} 85.8 & \secondc{} 75.0 & \secondc{} 80.0 & \secondc{} 54.5 & \secondc{} 38.9 & \secondc{} 81.5 & 78.8 & \firstc{} 42.9 & \firstc{} 100.0 & \firstc{} 66.7 & \firstc{} 95.0 & \firstc{} 90.0 & 0.0 & \secondc{} 67.6$\pm$0.1 \\ 
        \whline
    \end{tabular}
    }
\end{table*}

\begin{table}[t]
  \caption{\benchname{} evaluation of LLaVA, MM-ReAct and GPT-4V regarding each \textit{core VL capability}. For each column, the highest and second figures are highlighted by \setlength{\fboxsep}{0pt}\colorbox[RGB]{ 171, 235, 198 }{green} and \setlength{\fboxsep}{0pt}\colorbox[RGB]{ 253, 235, 208}{orange} colors. %
  }
  \label{tab:results_cap_gptv}
  \vspace{-3mm}
  \centering
  \scriptsize
  \setlength{\tabcolsep}{2pt}
    \begin{tabular}{lcccccccc}
    \whline
        Model & Rec & OCR & Know & Gen & Spat & Math & Total  \\ 
        \whline
        LLaVA-13B (LLaMA-2) 
        & 39.2 & 22.7 & 26.5 &  29.3 & 29.6 & 7.7 & 32.9$\pm$0.1 \\
        MM-ReAct-GPT-4 
        & 33.1 & 65.7 & 29.0 & 35.0 & 56.8 & \secondc{} 69.2 & 44.6$\pm$0.2 \\ 
        GPT-4V 
        & \firstc{} 67.5 & \secondc{} 68.3 & \firstc{} 56.2 & \firstc{} 60.7 & \secondc{} 69.4 & 58.6 & \firstc{} 67.7$\pm$0.3 \\ 
        GPT-4V-Turbo-detail:low 
        & 61.3 & 59.2 & \secondc{} 54.8 & \secondc{} 60.2 & 58.4 & 46.2 & 60.2$\pm$0.3 \\ 
        GPT-4V-Turbo-detail:high 
        & \secondc{} 62.9 & \firstc{} 75.9 & 53.7 & 57.3 & \firstc{} 76.8 & \firstc{} 69.5 & \secondc{} 67.6$\pm$0.1 \\ 
        \whline
    \end{tabular}
  \vspace{-4mm}
\end{table}

\vspace{-.05in}
\subsubsection{Regarding each capability integration}
\vspace{-.1in}
\myPara{Recognition, knowledge, and language generation} 
As shown in Appendix Table \ref{tab:samples1}(a), this capability integration can enable models to explain visual jokes. 
LLaMA-Adapter-v2-7B \cite{gao2023llama} is the best model in this capability integration. This may be attributed to its large scale of tuning data as shown in Table \ref{tab:model_details}. 
LLaVA-13B (LLaMA-2) and LLaVA-13B (V1.3, 336px) \cite{llava} are the other two outstanding models. Stronger language models and the tuning data in LLava may be the reason for the good performance.%

\myPara{Recognition (sole)} This category contains samples that only require recognition, as shown in Appendix Table \ref{tab:samples1}(b). 
InstructBLIP-14B and InstructBLIP-8B \cite{dai2023instructblip} achieve the best performance, which may result from the tuning data containing relevant datasets, like VQA~\cite{vqav2} and GQA~\cite{hudson2019gqa}.

\myPara{OCR and spatial awareness} For this integration, an example is shown in Appendix Table \ref{tab:samples1}(c).
MM-ReAct-GPT-4 \cite{yang2023mm} is the best method for this integration. Notably, MM-ReAct-GPT-4 has a significant improvement of over 40\% than MM-ReAct-GPT-3.5, indicating the importance of LLMs in integrating the OCR and location information.

\myPara{OCR, spatial awareness, and math} An example of this integration is shown in Appendix Table \ref{tab:samples2}(a), which requires reading the floor plan and conducting arithmetic. 
Compared with the above integration, this combination involves one more capability of math. The observation is similar to the integration of OCR and spatial awareness. MM-ReAct-GPT-4 \cite{yang2023mm} still achieves the best performance.

\myPara{Recognition and spatial awareness} Appendix Table \ref{tab:samples2}(b) shows an example for this integration. LLaVA-13B (V1.3, 336px) \cite{llava} performs best for this category. Compared with LLaVA-13B (LLaMA-2), LLaVA-13B (V1.3, 336px) obtains an improvement of 8.4\%, indicating the significant contribution of larger resolution of images. 

\myPara{OCR (sole)} This task requires OCR only, as shown in Appendix Table \ref{tab:samples2}(c). 
MM-ReAct-GPT-4 \cite{yang2023mm} has the best results for sole OCR due to an OCR tool from Azure API.  Notably, MM-ReAct-GPT-4 is much better than MM-ReAct-GPT-3.5 with an improvement of 23.0\%,  demonstrating the importance of language models in OCR.

\myPara{OCR and math} 
This integration enables reading text from real-world scenarios and solving math problems, as shown in Appendix Table \ref{tab:samples2}(d).  
MM-ReAct-GPT-4~\cite{yang2023mm} obtains the best performance in this capability integration, thanks to the speclized OCR and math tools used. %

\myPara{Other capability integrations} 9 other capability integrations are in long-tailed distribution, where MMReAct-GPT-4 achieves the best scores in 5 integrations out of 9. Their examples are shown in Appendix Tables \ref{tab:samples3}-\ref{tab:samples6}.

\subsection{Result discussion}
\subsubsection{Foundation models and tuning data}
\vspace{-1mm}
\label{sec:moduleana}
In this subsection, we discuss LMM modules and speculate how each component may affect the LMMs' capabilities in different aspects, evaluated by \benchname{}. We mainly consider the models based on open-sourced LLMs, \ie, Flan-T5~\cite{chung2022scaling}, LLaMA~\cite{touvron2023llama}, Vicuna~\cite{zheng2023judging}, and LLaMA-2~\cite{touvron2023llama2}.

\myPara{Vision} For the vision component, two models are popular in our evaluated end-to-end LMMs, \ie, CLIP-ViT/L14 \cite{radford2021learning} (428M) and EVA-ViT-G (1.13B). 
Determining a superior model is currently not possible due to the absence of a comprehensive ablation study~\cite{zeng2023matters}. However, it's noteworthy that, when paired with the same language model, Vicuna-7B, InstructBLIP-8B excels in recognition tasks, while LLaVA-7B works particularly well for OCR.

\myPara{Language} There is a notable trend indicating that superior language models (LLMs) typically yield better performance, such as comparing the 7B and 13B variants of different models, except for the outlier of InstructBLIP where the 8B version performs better than the 14B one. 

\myPara{Tuning data} Increasing the volume of data can enhance performance. An example is InstructBLIP-8B \cite{dai2023instructblip}, which utilizes more data from 26 publicly available datasets to tune the model and achieve higher scores than BLIP-2-12B. 

\vspace{-2mm}
\subsubsection{Comparison with GPT-4V(ision)}
\vspace{-2mm}
\label{sec:gpt4v}
We evaluate and benchmark the state-of-the-art LMM, GPT-4V(ison)~\cite{openai2023gpt4,gpt4v,gpt4vcontribution,gpt4vblog,yang2023dawn} on \benchname{}. In our queries to GPT-4V, we prepend the prompt with ``Generate a short and concise response to the following image text pair.'' The quantitative results are shown in Tables \ref{tab:results_cap_integration_gptv} and \ref{tab:results_cap_gptv}, and the qualitative results are expressed in Appendix Figures \ref{fig:gpt4v_1}-\ref{fig:gpt4v_4}. Remarkably, GPT-4V achieves a score of 67.7\%, surpassing both open-sourced LMMs~\cite{llava} and LLM-based multimodal agents~\cite{yang2023mm} by substantial margins. 

We aspire that the detailed per-category performance breakdown sheds light on potential avenues for enhancing model capabilities, thereby bridging the existing performance gap. To illustrate, integrating specialized tools within agent systems proves advantageous for specific functionalities like OCR and math. While other categories, such as recognition and language generation, would require enhancements in the core vision and language modules, respectively. Appendix Figures \ref{fig:gpt4v_1}-\ref{fig:gpt4v_4} offer an exhaustive analysis, highlighting representative success and failure instances of GPT-4V's performance.

\begin{table*}[t]
  \caption{Averaged absolute differences ($\overline{\Delta}$) between the evaluation scores of various LLM evaluators and those of human-annotated scores, on MM-ReAct-GPT4's results. A smaller discrepancy indicates a better agreement with the gold standard of human evaluation, indicating a better evaluator.
  }
  \label{tab:resutls_eval_diff}
  \vspace{-3mm}
\scriptsize
\setlength{\tabcolsep}{3pt}
  \centering
  \scalebox{0.8}{
    \begin{tabular}{l c c c c c c c c c c c}
    \whline
        \multirow{2}{*}{\makecell[c]{Model}} & \multirow{2}{*}{\makecell[c]{Keyword \\ matching}} & \multicolumn{10}{c}{LLM-based evaluation}  \\ 
        ~ & ~ & LLaMA-2-7B & LLaMA-2-13B & LLaMA-2-70B & Mistral-7B-v0.1 & Mistral-7B-Instruct-v0.2	& Mixtral-8x7B-Instruct-v0.1 & Gemini Pro & Claude 3 Opus & GPT-3.5 (turbo-0613) & GPT-4 (0613) \\
        \whline
        $\overline{\Delta}~(\downarrow)$ & 0.273 & 0.307 & 0.254 & 0.316 & 0.188 & 0.173 & 0.234 & 0.144 & 0.137 & 0.178 & \textbf{0.042} \\
        \whline
    \end{tabular}
    }
    \vspace{-2mm}
\end{table*}

\begin{table}[h]
  \caption{Averaged absolute differences ($\overline{\Delta}$) between the evaluation scores of combined LLM evaluators and those of human-annotated scores on MM-ReAct-GPT4's results. 
  }
  \label{tab:resutls_eval_diff_combined_llm}
  \vspace{-3mm}
\scriptsize
\setlength{\tabcolsep}{3pt}
  \centering
  \scalebox{1}{
    \begin{tabular}{l c }
    \whline
    Combined LLMs & $\overline{\Delta}$ \\
    \whline
    GPT-4 solely & \textbf{0.0423} \\
    GPT-4 + Llama-2-13b-chat & 0.1483 \\
    GPT-4 + Mistral-7B-Instruct-v0.2 & 0.1078 \\
    GPT-4 + Gemini Pro & 0.0933 \\
    GPT-4 + Claude 3 Opus & 0.0896 \\
    \makecell[l]{GPT-4 + Llama-2-13b-chat + \\ Mistral-7B-Instruct-v0.2 + Gemini Pro + Claude 3 Opus} & 0.1502 \\
    \whline
    \end{tabular}
    }
    \vspace{-2mm}
\end{table}

\subsection{Effectiveness analysis of LLM-based evaluation}
\label{sec:eval}
To verify the effectiveness of LLM-based evaluation for LMM predictions, we select the outputs from MMReAct-GPT-4 on 138 objective questions, which can be objectively annotated by humans. We compute the absolute value of the difference between the evaluator's output score and the human-annotated score on each sample. In addition to the default evaluator of GPT-4 (0613) in \benchname{}, we experiment with other LLMs. LLaMa-2 and Mixtral represent open-source LLMs, while GPT-4, Gemini and Claude are commercial close-sourced LLMs.

The average difference to the human scoring is reported in Table~\ref{tab:resutls_eval_diff}, represented as $\overline{\Delta}$.
With a maximum potential discrepancy of 1.0, the baseline evaluation method, keyword matching, results in a high difference of 0.273. This illustrates the unsuitability of keyword matching for \benchname{} when dealing with open-ended answers.
Among the LLaMA-2/Mistral series, LLaMA-2-13B/Mistral-7B-Instruct-v0.2 performs best, respectively. We notice that the performance of the largest LLaMA-2/Mistral is not satisfactory, which may be because their larger models are more creative and do not follow our few-shot prompt strictly. The three commercial LLMs perform better than open-sourced LLMs, while there is still a large gap between Gemini/Claude and GPT-4.

We also explore whether $\overline{\Delta}$ can be reduced when combining GPT-4 with other LLMs, and the results are reported in Table \ref{tab:resutls_eval_diff_combined_llm}. We find that GPT-4 without extra LLMs performs the best. This may be because other LLMs cannot match the grading accuracy of GPT-4. However, we believe this idea will work in the future when other LLMs become stronger. Therefore, \benchname{} uses GPT-4 (0613) to evaluate the LMM outputs.

\subsection{Effectiveness of few-shot examples for LLM prompt}

In this section, we explore the effectiveness of few-shot examples used in the prompt of LLM-based evaluator. We denote the seven grading examples in  Table \ref{tab:prompt} as (1) - (7) in order for ablation study. As Table \ref{tab:ablation_prompt} shows, using all seven examples together achieves the closest alignment with human evaluations (lowest $\overline{\Delta}$).

\begin{table*}[h]
  \caption{Ablation study of few-shot examples in prompt for LLM-based evaluator
  }
  \label{tab:ablation_prompt}
  \vspace{-3mm}
\scriptsize
\setlength{\tabcolsep}{3pt}
  \centering
  \scalebox{1}{
    \begin{tabular}{l c c}
    \whline
    Few-shot examples & Remarks & $\overline{\Delta}$ \\
    \whline
    None & No grading examples & 0.0630 \\
    (1) (5) & Add two examples with totally right/wrong short answer & 0.0625 \\
    (1) (2) (3) (5) & Add more two examples with partially right short answer & 0.0619 \\
    (1) (2) (3) (4) (5) & Add more one example with partially right and partially wrong answer & 0.0551 \\
    (1) (2) (3) (4) (5) (7) & Add more one example with fully right long answer & 0.0520 \\
    (1) (2) (3) (4) (5) (6) (7) & Add more one example with partially right long answer & \textbf{0.0423} \\
    \whline
    \end{tabular}
    }
    \vspace{-2mm}
\end{table*}

\subsection{Takeaway notes}
\label{sec:takeaway}
\vspace{-2mm}
We summarize the analyses and discussions as follows:
\begin{itemize}
    \item In the evaluation of integrated capabilities on \benchname{} (Sections~\ref{sec:result} and \ref{sec:gpt4v}), GPT-4V~\cite{gpt4v} outperforms existing open-sourced methods. 
    The tool-using approach, MM-ReAct-GPT-4 \cite{yang2023mm}, achieves the second-best performance with effective external tools. The pros and cons in different categories motivate future studies on tool-enhanced LMMs. 
    Among end-to-end LMMs, LLaVA-13B (LLaMA-2)/LLaVA-13B (V1.3, 336px)~\cite{llava} demonstrates the best performance on \benchname{}.

    \item Analysis of open-source LMMs (Section~\ref{sec:moduleana}) presents some uncertainty about which vision encoders are optimal for LMMs, based on current model comparisons. However, it is evident that stronger LLMs can boost the performance of LMMs.

    \item For open-ended evaluation (Section~\ref{sec:eval}), it is effective to use GPT-4 for evaluating the open-ended outputs of LMMs. The use of less powerful LLMs could result in more significant deviations from the gold standard of human evaluation results.

    \item Current top-performing methods, such as GPT-4V \cite{gpt4v}, only achieve scores of around 68\% on \benchname{}, where full score is 100\%. The gap signifies that further effort is necessary to enhance the performance of LMMs in terms of integrated capabilities, \eg, by developing stronger LMMs, finding better prompting techniques~\cite{yang2023dawn,yang2023setofmark}, and extending LMMs with external tools.
\end{itemize}

\section{Conclusion and Limitation}
In this paper, we have presented \benchname{}, a new benchmark designed to evaluate LMMs in terms of their integrated VL capabilities. We have constructed a new multimodal dataset that requires the integration of multiple VL capabilities to solve. To facilitate open-ended evaluation, we adopt an LLM-based evaluator to grade open-ended outputs from LMMs. We then evaluate various LMMs on \benchname{}, analyzing their results to provide insights into different LMM system paradigms and model designs. 
The evaluation reveals that even advanced models like GPT-4V only score around 68\% on \benchname{}, highlighting the ongoing need to enhance the integrated VL capabilities of LMMs. %

For the limitations of this work, firstly, since most popular LMMs only accept image-text input and output text, MM-Vet focuses on evaluating this type of modalities, not covering other modalities. Secondly, we propose to utilize LLMs to automatically grade LMM output results. However, as Table \ref{tab:resutls_eval_diff} shows, currently only GPT-4 can well match the human grades, so we set the evaluator’s LLM as GPT-4, which will bring GPT-4 usage fees. To help researchers from non-profit institutes save GPT-4 fees, we host an MM-Vet online evaluator \footnote{\href{https://huggingface.co/spaces/whyu/MM-Vet_Evaluator}{hf.co/spaces/whyu/MM-Vet\_Evaluator}} with our GPT-4 API key.

\section*{Acknowledgements}
This project is supported by 
the Ministry of Education, Singapore, under its Academic Research Fund Tier 2 (Award Number: MOE-T2EP20122-0006).
Weihao was partly supported by Snap Research Fellowship, Google TPU Research Cloud (TRC), and Google Cloud Research Credits program.

\section*{Impact Statement}
This paper presents a new benchmark for evaluating Large Multimodal Models (LMMs). The significance of this contribution extends in several directions. First, the newly developed MM-Vet benchmark assesses the capabilities of existing LMMs, establishing a solid groundwork for future advancements in this domain. The growing interest and demand for advanced LMM infrastructures underscore the timeliness and relevance of our work. 
Moreover, MM-Vet is pioneering in its approach to evaluating LMMs from integrated capabilities and leveraging LLMs for based open-ended scoring. These innovative evaluation strategies are poised to profoundly influence future benchmark development. By doing so, MM-Vet aims to not just enhance the field of LMMs but also to play a pivotal role in driving societal changes through the application of advanced foundational models. In the development of MM-Vet, ethical considerations have been paramount. We have rigorously verified the benchmark sample and evaluation criteria to ensure they align with high ethical standards. This diligence is vital in ensuring that the advancements in LMMs are responsible and beneficial for society at large.

\bibliography{main}
\bibliographystyle{icml2024}

\newpage
\appendix
\onecolumn
\section{Model details}

The details of the models we evaluated are shown in the Table \ref{tab:model_details}.

\section{Comparison with Bard}
\vspace{-1mm}
\label{sec:bard}
Bard~\cite{bard} is another representative closed-source commercial LMM system. One problem in evaluation is that Bard refuses to process images containing people faces.
To conduct a fair comparison, we constructed a subset of \benchname{} with 168 samples that Bard could process, henceforth referred to as the Bard set.
The results on the Bard set are shown in Tables \ref{tab:results_cap_integration_bard_set} and \ref{tab:resutls_cap_bard_set}. 
Bard achieves the highest scores in three out of six capabilities, seven out of fifteen capability integrations, and holds the highest overall score (53.5\%). MM-ReAct-GPT-4~\cite{yang2023mm} outperforms in the remaining three out of six capabilities, and tops the chart in nine out of the fifteen capability integrations. Particularly, MM-ReAct performs better in OCR, spatial awareness, and math capabilities, indicating the potential benefit of having specialized external tools, even when working with state-of-the-art LMMs. 
When considering open-sourced end-to-end models such as LLaVA, there is still a considerable gap. %

\begin{table}[h!]
  \caption{\benchname{} (Bard set) evaluation results on various LMMs regarding each \textit{core VL capability}. For each column, the highest and second figures are highlighted by \setlength{\fboxsep}{0pt}\colorbox[RGB]{ 171, 235, 198 }{green} and \setlength{\fboxsep}{0pt}\colorbox[RGB]{ 253, 235, 208}{orange} colors. %
  }
  \label{tab:resutls_cap_bard_set}
  \centering
  \setlength{\tabcolsep}{2pt}
    \begin{tabular}{lcccccccc}
    \whline
        Model & Rec & OCR & Know & Gen & Spat & Math & Total  \\ 
        \whline
        LLaVA-13B (LLaMA-2) \cite{llava} & 37.8 & 22.9 & 22.4 & 27.6 & 27.2 & 8.0 & 30.3$\pm$0.1 \\
        LLaVA-13B (V1.3, 336px) \cite{llava} & \secondc{} 39.4 & 22.3 & 22.7 & 24.6 & 30.6 & 11.6 & 31.5$\pm$0.1 \\
        \hline 
        MM-ReAct-GPT-3.5~\cite{yang2023mm} & 22.3 & 31.4 & 15.6 & 16.6 & 32.9 & 24.0 & 27.6$\pm$0.2 \\ 
        MM-ReAct-GPT-4~\cite{yang2023mm} & 34.3 & \firstc{} 66.3 & \secondc{} 25.6 & \secondc{} 36.6 & \firstc{} 60.6 & \firstc{} 72.0 & \secondc{} 48.1$\pm$0.2 \\ 
        \hline
        Bard  \cite{bard} & \firstc{} 56.2 & \secondc{} 52.5 & \firstc{} 50.9 & \firstc{} 61.0 & \secondc{} 52.0 & \secondc{} 39.6 & \firstc{} 53.5$\pm$0.2 \\
        \whline
    \end{tabular}
\end{table}

\begin{table*}[h!]
  \caption{\benchname{} (Bard set) evaluation results on various LMMs regarding each \textit{capability integration}. For each column, the highest and second highest figures are highlighted by \setlength{\fboxsep}{0pt}\colorbox[RGB]{ 171, 235, 198 }{green} and \setlength{\fboxsep}{0pt}\colorbox[RGB]{ 253, 235, 208}{orange} colors. Numbers are presented in \% with a full score of 100\%.
  }
  \label{tab:results_cap_integration_bard_set}
  \scriptsize
  \vspace{-0.12in}
  \setlength{\tabcolsep}{2pt}
  \centering
    \scalebox{1}{
    \begin{tabular}{lcccccccccccccccccc}
    \hline
        \multirow{4}{*}{\makecell[c]{Model}} & \multirow{4}{*}{\makecell[c]{~ \\ Rec \\ Know \\ Gen}} & \multirow{4}{*}{\makecell[c]{~ \\ ~ \\ ~ \\ Rec }} & \multirow{4}{*}{\makecell[c]{~ \\ ~ \\ OCR \\ Spat}} & \multirow{4}{*}{\makecell[c]{~ \\ OCR \\ Spat \\ Math}} & \multirow{4}{*}{\makecell[c]{~ \\ ~ \\ Rec \\ Spat}} & \multirow{4}{*}{\makecell[c]{~ \\ ~ \\ ~ \\ OCR}} & \multirow{4}{*}{\makecell[c]{~ \\ ~ \\ OCR \\ Math}} & \multirow{4}{*}{\makecell[c]{~ \\ ~ \\ Rec \\ Know}} & \multirow{4}{*}{\makecell[c]{Rec \\ OCR \\ Know \\ Gen }}  & \multirow{4}{*}{\makecell[c]{Rec \\ OCR \\ Gen \\ Spat }} & \multirow{4}{*}{\makecell[c]{~ \\ Rec \\ OCR \\ Spat }} & \multirow{4}{*}{\makecell[c]{~ \\ ~ \\ Rec \\ OCR }} & \multirow{4}{*}{\makecell[c]{~ \\ OCR \\ Know \\ Spat }} & \multirow{4}{*}{\makecell[c]{~ \\ Rec \\ Know \\ Spat}}  & \multirow{4}{*}{\makecell[c]{~ \\ OCR \\ Gen \\ Spat }} & \multirow{4}{*}{\makecell[c]{Rec \\ OCR \\ Spat \\ Math }} & \multirow{4}{*}{\makecell[c]{Total}} \\
        ~ \\
        ~ \\
        ~ \\
        \whline
        Vicuna-13B (LLaMA-2) \cite{llava} & \secondc{} 26.6 & 55.2 & 18.8 & 14.3 & \secondc{} 57.1 & 39.5 & 0.0 & \secondc{} 20.0 & 1.3 & 56.8 & \firstc{} 28.6 & \firstc{} 50.0 & 33.3 & 0.0 & 8.0 & -- & 30.3$\pm$0.1 \\
        Vicuna-13B (V1.3, 336px) \cite{llava} & 21.9 & \secondc{} 59.0 & 22.9 & 14.3 & \firstc{} 85.7 & 25.5 & 8.2 & \secondc{} 20.0 & 15.0 & 49.3 & 14.3 & \firstc{} 50.0 & 33.3 & \firstc{} 50.0 & 2.0 & -- & 31.5$\pm$0.1 \\
        \hline
        MM-ReAct-GPT-3.5~\cite{yang2023mm} & 11.3 & 38.8 & 31.2 & 35.7 & 28.6 & 56.4 & 9.1 & \secondc{} 20.0 & 0.0 & 47.8 & 0.0 & 25.0 & \firstc{} 100.0 & 0.0 & 35.0 & -- & 27.6$\pm$0.2 \\ 
        MM-ReAct-GPT-4~\cite{yang2023mm} & 17.0 & 35.2 & \firstc{} 70.8 & \firstc{} 78.6 & 28.6 & \firstc{} 81.5 & \firstc{} 63.6 & \firstc{} 40.0 & \secondc{} 68.3 & \firstc{} 88.0 & 14.3 & \firstc{} 50.0 & 0.0 & \firstc{} 50.0 & \firstc{} 80.0 & -- & \secondc{} 48.1$\pm$0.2 \\ 
        \hline
        Bard \cite{bard} & \firstc{} 52.3 & \firstc{} 70.3 & \secondc{} 45.2 & \secondc{} 56.4 & 42.9 & \secondc{} 70.2 & \secondc{} 18.2 & 0.0 & \firstc{} 77.7 & \secondc{} 81.5 & \firstc{} 28.6 & \firstc{} 50.0 & \secondc{} 66.7 & \firstc{} 50.0 & \firstc{} 80.0 & -- & \firstc{} 53.5$\pm$0.2 \\
        \whline
    \end{tabular}
    }

\end{table*}

\begin{sidewaystable}[h]
  \caption{Summary of the evaluated LMMs in this report. We consider both the end-to-end tuned models (\ie, OpenFlamingo \cite{alayrac2022flamingo, anas_awadalla_2023_7733589, awadalla2023openflamingo}, BLIP-2 \cite{li2023blip}, LLaVA \cite{llava}, MiniGPT-4 \cite{zhu2023minigpt}, LLaMA-Adapter v2 \cite{gao2023llama}, Otter \cite{li2023otter}, and InstructBLIP \cite{dai2023instructblip}), and the LLM-tool-using systems (\ie, MM-ReAct \cite{yang2023mm} and Transformers Agent \cite{transformers_agent}). }
  \centering
  \vspace{-0.12in}
  \scriptsize
  \setlength{\tabcolsep}{6pt}
  \scalebox{0.75}{
    \begin{tabular}{l|lll|l|c}
    \whline
        \multirow{2}{*}{\makecell[c]{Method}} & \multicolumn{3}{c|}{Initial models}  &  \makecell*[c]{\multirow{2}{*}{Tuning data}} & \multirow{2}{*}{\makecell[c]{Total params}} \\ 
        ~ & \makecell[c]{Vision} & \makecell[c]{Language} & \makecell[c]{Other} & \\
        \whline
        \multirow{2}{*}{\makecell[l]{OpenFlamingo-9B \\ \cite{alayrac2022flamingo, anas_awadalla_2023_7733589, awadalla2023openflamingo}}} & \multirow{2}{*}{\makecell[l]{CLIP ViT-L/14 \cite{radford2021learning}}} & \multirow{2}{*}{\makecell[l]{MPT-7B \cite{MPT}}} & \multirow{2}{*}{\makecell[l]{\makecell[c]{--}}} &  \multirow{2}{*}{\makecell[l]{Multimodal C4 \cite{zhu2023multimodal}}} & \multirow{2}{*}{\makecell[c]{9B}} \\ 
        ~ & ~ & ~ & ~ & ~ & ~ \\
        \hline
        \multirow{8}{*}{\makecell[c]{BLIP-2-12B \cite{li2023blip}}}  & \multirow{8}{*}{\makecell[c]{EVA-ViT-G \cite{fang2023eva}}} & \multirow{8}{*}{\makecell[c]{Flan-T5-XXL \cite{chung2022scaling}}} & \makecell*[c]{\multirow{8}{*}{--}} & \multirow{8}{*}{\makecell[l]{1. COCO \cite{lin2014microsoft}; \\ 2. Visual Genome \cite{krishna2017visual}; \\ 3. CC3M \cite{sharma2018conceptual}; \\ 4. CC12M \cite{changpinyo2021conceptual}; \\ 5. SBU \cite{ordonez2011im2text}; \\ 6. 115M images from the LAION-400M \cite{schuhmann2021laion}. \\ (CapFilt \cite{li2022blip} is used to create synthetic \\ captions for the web images)}}  & \multirow{5}{*}{\makecell[c]{12B}} \\ 
        ~ & ~ & ~ & ~ & ~ & ~ \\
        ~ & ~ & ~ & ~ & ~ & ~ \\
        ~ & ~ & ~ & ~ & ~ & ~ \\
        ~ & ~ & ~ & ~ & ~ & ~ \\
        ~ & ~ & ~ & ~ & ~ & ~ \\
        ~ & ~ & ~ & ~ & ~ & ~ \\
        ~ & ~ & ~ & ~ & ~ & ~ \\
        \hline
        \multirow{3}{*}{\makecell[c]{\\ LLaVA-7B \cite{llava} \\ \\  LLaVA-13B \cite{llava} }}&  \multirow{3}{*}{\makecell[c]{CLIP ViT-L/14 \cite{radford2021learning}}} & \multirow{3}{*}{\makecell[c]{Vicuna-7B \cite{zheng2023judging} \\ \\ Vicuna-13B \cite{zheng2023judging}}} & \makecell*[c]{\multirow{3}{*}{--}} & \multirow{3}{*}{\makecell[l]{1. CC3M \cite{sharma2018conceptual} \\ Concept-balanced 595K \cite{llava}; \\
        2. LLaVA-Instruct-158K \cite{llava}.
        }}  & \multirow{3}{*}{\makecell[c]{7B \\ 13B}} \\ 
        ~ & ~ & ~ & ~ & ~ & ~ \\
        ~ & ~ & ~ & ~ & ~ & ~ \\
        \hdashline
        LLaVA-7B (LLaMA-2) \cite{llava} &  \multirow{2}{*}{\makecell[c]{CLIP ViT-L/14 \cite{radford2021learning}}} & LLaMA-2-7B-Chat \cite{touvron2023llama2} & \makecell*[c]{\multirow{3}{*}{--}} & \multirow{3}{*}{\makecell[l]{1. LAION /CC/SBU BLIP-Caption \\ Concept-balanced 558K \cite{llava}; \\
        2. LLaVA-Instruct-80K \cite{llava}.
        }}  & 7B \\ 
        LLaVA-13B (LLaMA-2) \cite{llava} & ~ & LLaMA-2-13B-Chat \cite{touvron2023llama2} & ~ & ~ & 13B \\
        LLaVA-13B (V1.3, 336px) \cite{llava} & CLIP ViT-L/336px \cite{radford2021learning} & Vicuna-13B-v1.3 \cite{zheng2023judging} & ~ & ~ & 13B \\
        \hline
        \multirow{5}{*}{\makecell[l]{\\MiniGPT-4-8B \cite{zhu2023minigpt} \\ \\ MiniGPT-4-14B \cite{zhu2023minigpt} \\ \\ }} & \multirow{5}{*}{\makecell[l]{EVA-ViT-G \cite{fang2023eva}}} & \multirow{5}{*}{\makecell[l]{\\ Vicuna-7B \cite{zheng2023judging} \\ \\  Vicuna-13B \cite{zheng2023judging} \\ \\ }} & \multirow{5}{*}{\makecell[c]{BLIP-2's Q-Former \cite{li2023blip}}} & \multirow{5}{*}{\makecell[l]{1. CC3M \cite{sharma2018conceptual}; \\ 2. CC12M \cite{changpinyo2021conceptual}; \\ 3. SBU \cite{ordonez2011im2text}; \\ 4. LAION-400M \cite{schuhmann2021laion} \\
        5. Proposed 3,500 aligned image-text pairs \cite{zhu2023minigpt}.}} & \multirow{5}{*}{\makecell[c]{\\ 8B \\ \\ 14B \\ \\ }} \\ 
        ~ & ~ & ~ & ~ & ~ & ~ \\
        ~ & ~ & ~ & ~ & ~ & ~ \\
        ~ & ~ & ~ & ~ & ~ & ~ \\
        ~ & ~ & ~ & ~ & ~ & ~ \\
        \hline
        \multirow{8}{*}{\makecell[c]{LLaMA-Adapter v2-7B \cite{gao2023llama}}} & \multirow{8}{*}{\makecell[c]{CLIP ViT-L/14 \cite{radford2021learning}}} & \multirow{8}{*}{\makecell[c]{LLaMA-7B \cite{touvron2023llama}}} & \makecell*[c]{\multirow{8}{*}{--}} & \multirow{8}{*}{\makecell[l]{1. LAION-400M \cite{schuhmann2021laion}; \\ 2. COYO-700M \cite{kakaobrain2022coyo-700m}; \\ 3. Multimodal C4 \cite{zhu2023multimodal}; \\ 4. SBU \cite{ordonez2011im2text}; \\ 5. CC12M \cite{changpinyo2021conceptual}; \\ 6. COCO \cite{lin2014microsoft}; \\ 7. GPT-4-LLM \cite{peng2023instruction}; \\ 8. Tuning data of LLaVA \cite{llava}}} & \multirow{8}{*}{\makecell[c]{7B}} \\ 
        ~ & ~ & ~ & ~ & ~ & ~ \\
        ~ & ~ & ~ & ~ & ~ & ~ \\
        ~ & ~ & ~ & ~ & ~ & ~ \\
        ~ & ~ & ~ & ~ & ~ & ~ \\
        ~ & ~ & ~ & ~ & ~ & ~ \\
        ~ & ~ & ~ & ~ & ~ & ~ \\
        ~ & ~ & ~ & ~ & ~ & ~ \\
        \hline
        \multirow{3}{*}{\makecell[c]{Otter-9B \cite{li2023otter}}} &  \multirow{3}{*}{\makecell[l]{CLIP ViT-L/14 \cite{radford2021learning}}} & \multirow{3}{*}{\makecell[c]{MPT-7B \cite{MPT}}} & \multirow{3}{*}{\makecell[l]{OpenFlamingo-9B's \\ 1. Perceiver
Resampler; \\ 2. GATED XATTN-DENSE}} &  \multirow{3}{*}{\makecell[l]{MIMIC-IT \cite{li2023mimic}}} & \multirow{3}{*}{\makecell[c]{9B}}  \\ 
        ~ & ~ & ~ & ~ & ~ & ~ \\
        ~ & ~ & ~ & ~ & ~ & ~ \\
        \hline
        InstructBLIP-8B \cite{dai2023instructblip} & \multirow{3}{*}{\makecell[c]{EVA-ViT-G \cite{fang2023eva}}} & Vicuna-7B \cite{zheng2023judging} & \multirow{3}{*}{\makecell[l]{BLIP-2's Q-Former \cite{li2023blip}}} & \multirow{3}{*}{\makecell[l]{1. Tuning data of BLIP-2 \cite{li2023blip}; \\
        2. 26 publicly available datasets \\ (transformed into instruction tuning format).}}  & 8B \\ 
        ~ & ~ & ~ & ~ & ~ & ~ \\
        InstructBLIP-14B \cite{dai2023instructblip} & ~ & Vicuna-13B \cite{zheng2023judging} & ~ & ~ & 14B \\
        \hline
        \multirow{7}{*}{\makecell[c]{Transformers Agent \\ (GPT-4 as agent) \cite{transformers_agent}}} & \makecell*[c]{\multirow{7}{*}{--}} & \multirow{7}{*}{\makecell[l]{1. GPT-4 \cite{openai2023gpt4}; \\ 2. Flan-T5 \cite{chung2022scaling}; \\ 3. BART \cite{lewis2019bart} }} & \multirow{7}{*}{\makecell[l]{1. Donut \cite{kim2022ocr}; \\ 2. BLIP \cite{li2022blip}; \\ 3. ViLT \cite{kim2021vilt}; \\ 4.  CLIPSeg \cite{luddecke2022image} \\ 5. Whisper \cite{radford2023robust}; \\ 6. SpeechT5 \cite{ao2021speecht5}; \\ 7. NLLB \cite{costa2022no}}} & \makecell*[c]{\multirow{7}{*}{None}} & \makecell*[c]{\multirow{7}{*}{Not clear}} \\
        ~ & ~ & ~ & ~ & ~ & ~ \\
        ~ & ~ & ~ & ~ & ~ & ~ \\
        ~ & ~ & ~ & ~ & ~ & ~ \\
        ~ & ~ & ~ & ~ & ~ & ~ \\
        ~ & ~ & ~ & ~ & ~ & ~ \\
        ~ & ~ & ~ & ~ & ~ & ~ \\
        \hline
        \multirow{6}{*}{\makecell[l]{MM-ReAct-GPT-3.5 \cite{yang2023mm} \\ ~ \\ MM-ReAct-GPT-4 \cite{yang2023mm} }} & \makecell*[c]{\multirow{6}{*}{--}} & \multirow{6}{*}{\makecell[l]{GPT-3.5 \cite{ouyang2022training} \\ ~ \\ GPT-4 \cite{openai2023gpt4}}} & \multirow{6}{*}{\makecell[l]{1. Azure Cognitive Services APIs \cite{azure_cognition_api} \\
        for image captioning, \\ image tagging, dense
captioning, \\ OCR and specialized recognition \\ on celebrities, receipts, \etc \\ 2. Bing search; 3. PAL \cite{gao2022pal}}} & \makecell*[c]{\multirow{6}{*}{None}} & \makecell*[c]{\multirow{6}{*}{Not clear}} \\
        ~ & ~ & ~ & ~ & ~ & ~ \\
        ~ & ~ & ~ & ~ & ~ & ~ \\
        ~ & ~ & ~ & ~ & ~ & ~ \\
        ~ & ~ & ~ & ~ & ~ & ~ \\
        ~ & ~ & ~ & ~ & ~ & ~ \\
        \whline 
    \end{tabular}
    }
  \label{tab:model_details}
  \vspace{-0.2in}
\end{sidewaystable}

\section{Examples of capability integrations}
We show examples of capability integrations in Tables \ref{tab:samples1}-\ref{tab:samples6}. More examples are shown on \url{https://github.com/yuweihao/MM-Vet/blob/main/README_more_samples.md}.

\begin{table*}[h!]
\caption{Three samples requiring different capability integrations. }
\label{tab:samples1}  
\centering  
\scalebox{0.9}{
\begin{tabular}{r p{11.5cm} }
\toprule
\textbf{(a)} \\
&  \includegraphics[height=4.5cm]{figures/examples/joke_8.jpg} \\
\textbf{Q:} & What occasions would someone use this meme? \\
\textbf{GT:} & This meme, commonly known as "Screaming Panda," is typically used to express shock, surprise, or fear. It could be used in response to a startling or unexpected event, or to convey a sense of panic or alarm. Some possible occasions where someone might use this meme include: \\
~ & ~ \\
~ &  - Reacting to a jump scare in a horror movie \\
~ & - Responding to a surprising plot twist in a TV show or book \\
~ & - Expressing shock at a news headline or current event\\
~ & - Conveying fear or anxiety about an upcoming deadline or exam \\
~ & - Showing surprise at an unexpected outcome in a sports game or other competition. \\
\textbf{Required capabilities:} & Recognition, knowledge, language generation \\ 
\midrule
\textbf{(b)} \\
~ & \includegraphics[height=4.5cm]{figures/examples/recognition_22.jpg} \\
\textbf{Q:} & How many tomatoes are there? \\
\textbf{GT:} & 5 \\
\textbf{Required capabilities:} & Recognition \\
\midrule
\textbf{(c)} \\
~ & \includegraphics[height=4.5cm]{figures/examples/spatial_1.jpg} \\
\textbf{Q:} &  What is located to the right of the shampoo?\\
\textbf{GT:} & conditioner \\
\textbf{Required capabilities:} & OCR, spatial awareness \\
\bottomrule
\end{tabular}
}

\end{table*}

\begin{table*}
\caption{Four samples requiring different capability integrations. }
\centering  
\scalebox{0.8}{
\begin{tabular}{r p{11.5cm} }
\toprule
\textbf{(a)} \\
&  \includegraphics[height=4.5cm]{figures/examples/document_5.jpg} \\
\textbf{Q:} & Which room is bigger, the double garage or the living room? \\
\textbf{GT:} & double garage \\
\textbf{Required capabilities:} & OCR, spatial awareness, math \\ 
\midrule
\textbf{(b)} \\
~ & \includegraphics[height=4.5cm]{figures/examples/reg_spat.jpg} \\
\textbf{Q:} & On the right desk, what is to the left of the laptop?  \\
\textbf{GT:} & table lamp <OR> desk lamp \\
\textbf{Required capabilities:} & Recognition, spatial awareness \\
\midrule
\textbf{(c)} \\
~ & \includegraphics[height=4.5cm]{figures/examples/ocr_3.jpg} \\
\textbf{Q:} &  What are all the scene text in the image? \\
\textbf{GT:} & 5:30PM <AND> 88\% <AND> Mario Kart 8 Deluxe <AND> MARIO KART 8 DELUXE <AND> SUPER MARIO ODYSSEY <AND> THE LEGEND OF ZELDA <AND> BREATH OF WILD <AND> Options <AND> Start \\
\textbf{Required capabilities:} & OCR \\
\midrule
\textbf{(d)} \\
~ & \includegraphics[height=4.5cm]{figures/examples/math_6.png} \\
\textbf{Q:} &  How many gallons of supreme gasoline can I get with \$50? \\
\textbf{GT:} & 13.6 <OR> 13.7 \\
\textbf{Required capabilities:} & OCR, math \\

\bottomrule
\end{tabular}
}
\label{tab:samples2}  
\end{table*}

\begin{table*}
\caption{Two samples requiring different capability integrations. }
\centering  
\scalebox{1}{
\begin{tabular}{r p{11.5cm} }
\toprule
\textbf{(a)} \\
&  \includegraphics[height=4.5cm]{figures/examples/news_3.png} \\
\textbf{Q:} & In which country was this photo taken? \\
\textbf{GT:} & Australia \\
\textbf{Required capabilities:} & Recognition, knowledge \\ 
\midrule
\textbf{(b)} \\
~ & \includegraphics[height=4.5cm]{figures/examples/joke_0.jpg} \\
\textbf{Q:} & Can you explain this meme?  \\
\textbf{GT:} & This meme is a humorous take on procrastination and the tendency to delay tasks until a specific time. The person in the meme plans to do something at 8 o'clock, but when they miss that deadline by a few minutes, they decide to wait until 9 o'clock instead. The image of Kermit the Frog lying in bed represents the person's laziness and lack of motivation to complete the task. \\
\textbf{Required capabilities:} & Recognition, OCR, knowledge, language generation \\
\bottomrule
\end{tabular}
}
\label{tab:samples3}  
\end{table*}

\begin{table*}
\caption{Two samples requiring different capability integrations. }
\centering  
\scalebox{1}{
\begin{tabular}{r p{11.5cm} }
\toprule
\textbf{(a)} \\
&  \includegraphics[height=4.5cm]{figures/examples/ielts_0.png} \\
\textbf{Q:} & The graph below shows the long-term international migration, UK, 1999-2008. \\
~ &  Summarize the information by selecting and reporting the main features, and make comparisons where relevant. \\
~ & You should write at least 150 words. \\
\textbf{GT:} & The chart gives information about UK immigration, emigration and net migration between 1999 and 2008. \\
~ & ~ \\
~ &  Both immigration and emigration rates rose over the period shown, but the figures for immigration were significantly higher. Net migration peaked in 2004 and 2007. \\
~ & ~ \\
~ & In 1999, over 450,000 people came to live in the UK, while the number of people who emigrated stood at just under 300,000. The figure for net migration was around 160,000, and it remained at a similar level until 2003. From 1999 to 2004, the immigration rate rose by nearly 150,000 people, but there was a much smaller rise in emigration. Net migration peaked at almost 250,000 people in 2004. \\
~ & ~ \\
~ & After 2004, the rate of immigration remained high, but the number of people emigrating fluctuated. Emigration fell suddenly in 2007, before peaking at about 420,000 people in 2008. As a result, the net migration figure rose to around 240,000 in 2007, but fell back to around 160,000 in 2008. \\
\textbf{Required capabilities:} & Recognition, OCR, language generation, spatial awareness  \\ 
\midrule
\textbf{(b)} \\
~ & \includegraphics[height=4.5cm]{figures/examples/spatial_3.jpg} \\
\textbf{Q:} & Which car is on the parking spot 33?  \\
\textbf{GT:} & no <OR> empty \\
\textbf{Required capabilities:} & Recognition, OCR, spatial awareness \\

\bottomrule
\end{tabular}
}
\label{tab:samples4}  
\end{table*}

\begin{table*}
\caption{Three samples requiring different capability integrations. }
\centering  
\scalebox{1}{
\begin{tabular}{r p{11.5cm} }
\toprule
\textbf{(a)} \\
&  \includegraphics[height=4.5cm]{figures/examples/recognition_4.png} \\
\textbf{Q:} & Is this apple organic? \\
\textbf{GT:} & yes \\
\textbf{Required capabilities:} & Recognition, OCR \\
\midrule
\textbf{(b)} \\
&  \includegraphics[height=4.5cm]{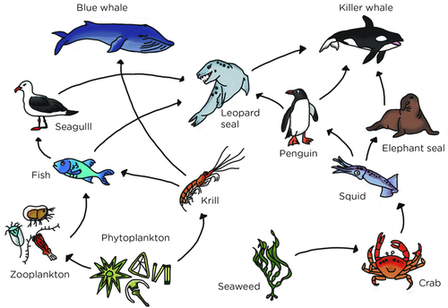} \\
\textbf{Q:} & Which are producers in this food web? \\
\textbf{GT:} & Phytoplankton <AND> Seaweed \\
\textbf{Required capabilities:} & OCR, knowledge, spatial awareness \\
\midrule
\textbf{(c)} \\
&  \includegraphics[height=4.5cm]{figures/examples/spatial_9.png} \\
\textbf{Q:} & Does the person bigger than the car? \\
\textbf{GT:} & no \\
\textbf{Required capabilities:} & Recognition, knowledge, spatial awareness \\
\bottomrule
\end{tabular}
}
\label{tab:samples5}  
\end{table*}

\begin{table*}
\caption{Two samples requiring different capability integrations. }
\label{tab:samples6}  
\centering  
\scalebox{1}{
\begin{tabular}{r p{11.5cm} }
\toprule
\textbf{(a)} \\
&  \includegraphics[height=4.5cm]{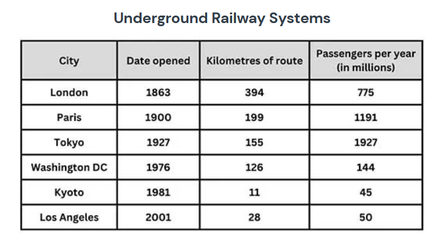} \\
\textbf{Q:} & The table below gives information about the underground railway systems in six cities. \\
~ & ~ \\
~ & Summarise the information by selecting and reporting the main features, and make comparisons where relevant. \\
~ & ~ \\
~ & You should write at least 150 words. \\
\textbf{GT:} & The table shows data about the underground rail networks in six major cities. \\
~ & ~ \\
~ & The table compares the six networks in terms of their age, size and the number of people who use them each year. It is clear that the three oldest underground systems are larger and serve significantly more passengers than the newer systems.\\
~ & ~ \\
~ & The London underground is the oldest system, having opened in 1863. It is also the largest system, with 394 kilometres of route. The second largest system, in Paris, is only about half the size of the London underground, with 199 kilometres of route. However, it serves more people per year. While only third in terms of size, the Tokyo system is easily the most used, with 1927 million passengers per year. \\
~ & ~ \\
~ & Of the three newer networks, the Washington DC underground is the most extensive, with 126 kilometres of route, compared to only 11 kilometres and 28 kilometres for the Kyoto and Los Angeles systems. The Los Angeles network is the newest, having opened in 2001, while the Kyoto network is the smallest and serves only 45 million passengers per year. \\
\textbf{Required capabilities:} & OCR, language generation, spatial awareness \\
\midrule
\textbf{(b)} \\
&  \includegraphics[height=4.5cm]{figures/examples/math_5.png} \\
\textbf{Q:} & What will the girl on the right write on the board? \\
\textbf{GT:} & 14 \\
\textbf{Required capabilities:} & Recognition, OCR, spatial awareness, math \\
\bottomrule
\end{tabular}
}
\end{table*}

\FloatBarrier
\newpage
\section{GPT-4V prediction examples}

\begin{figure*}[h!]
\centering
\centerline{\includegraphics[width=0.95\textwidth]{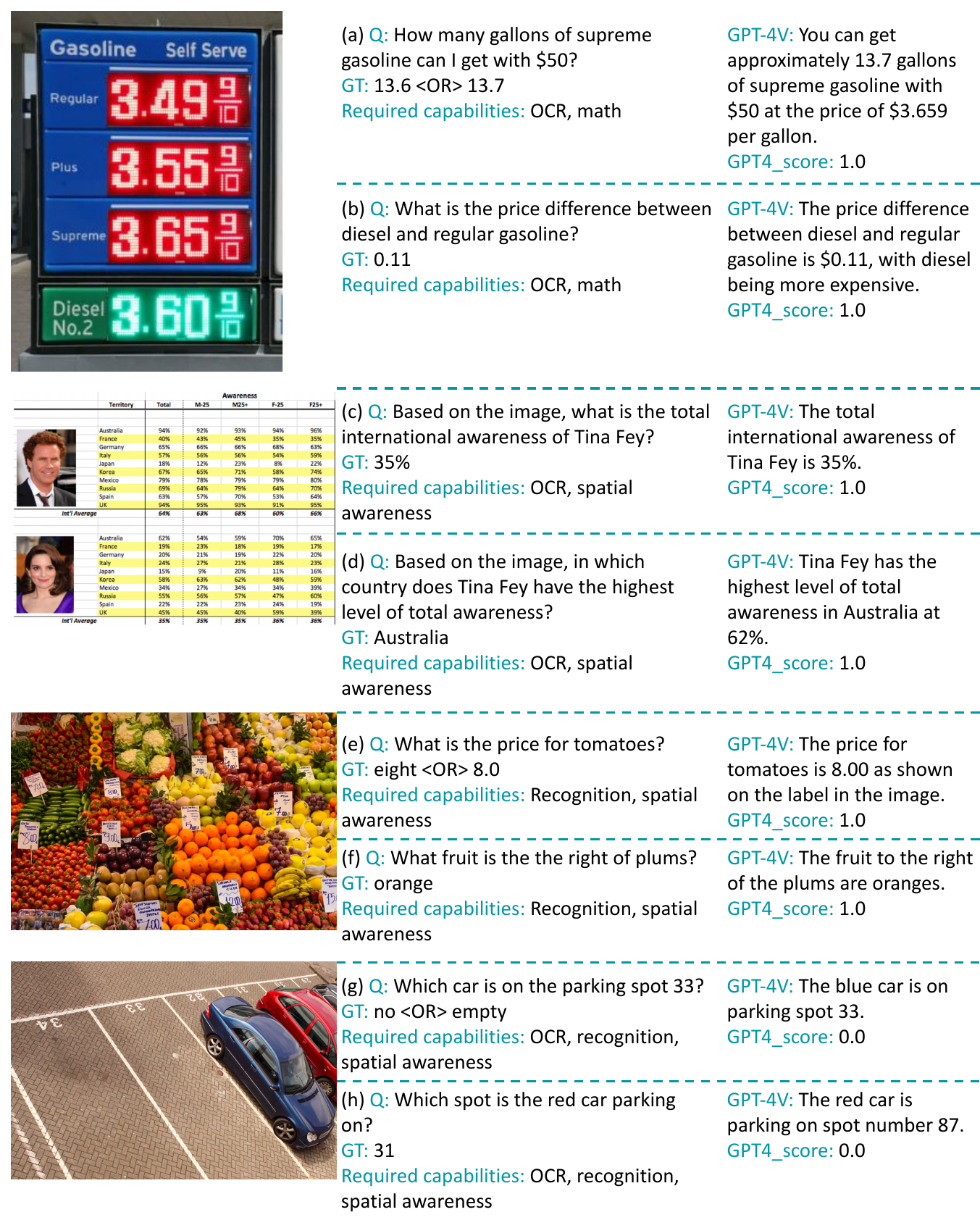}}
\caption{GPT-4V representative success and failure cases.
}
\label{fig:gpt4v_1}
\end{figure*}
\begin{figure*}[h!]
\centering
\centerline{\includegraphics[width=0.9\textwidth]{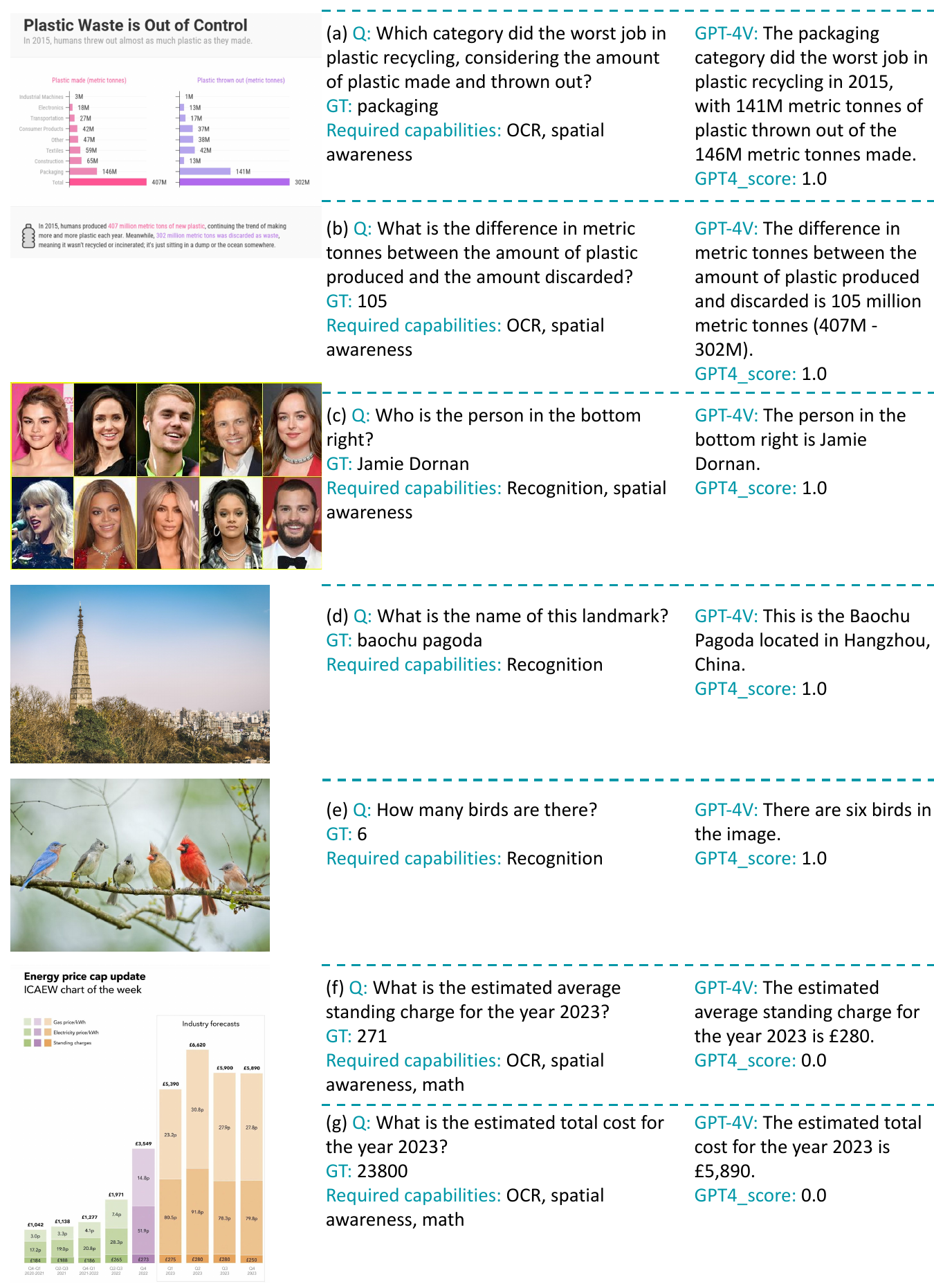}}
\caption{GPT-4V representative success and failure cases.
}
\label{fig:gpt4v_2}
\end{figure*}
\begin{figure*}[h!]
\centering
\centerline{\includegraphics[width=0.85\textwidth]{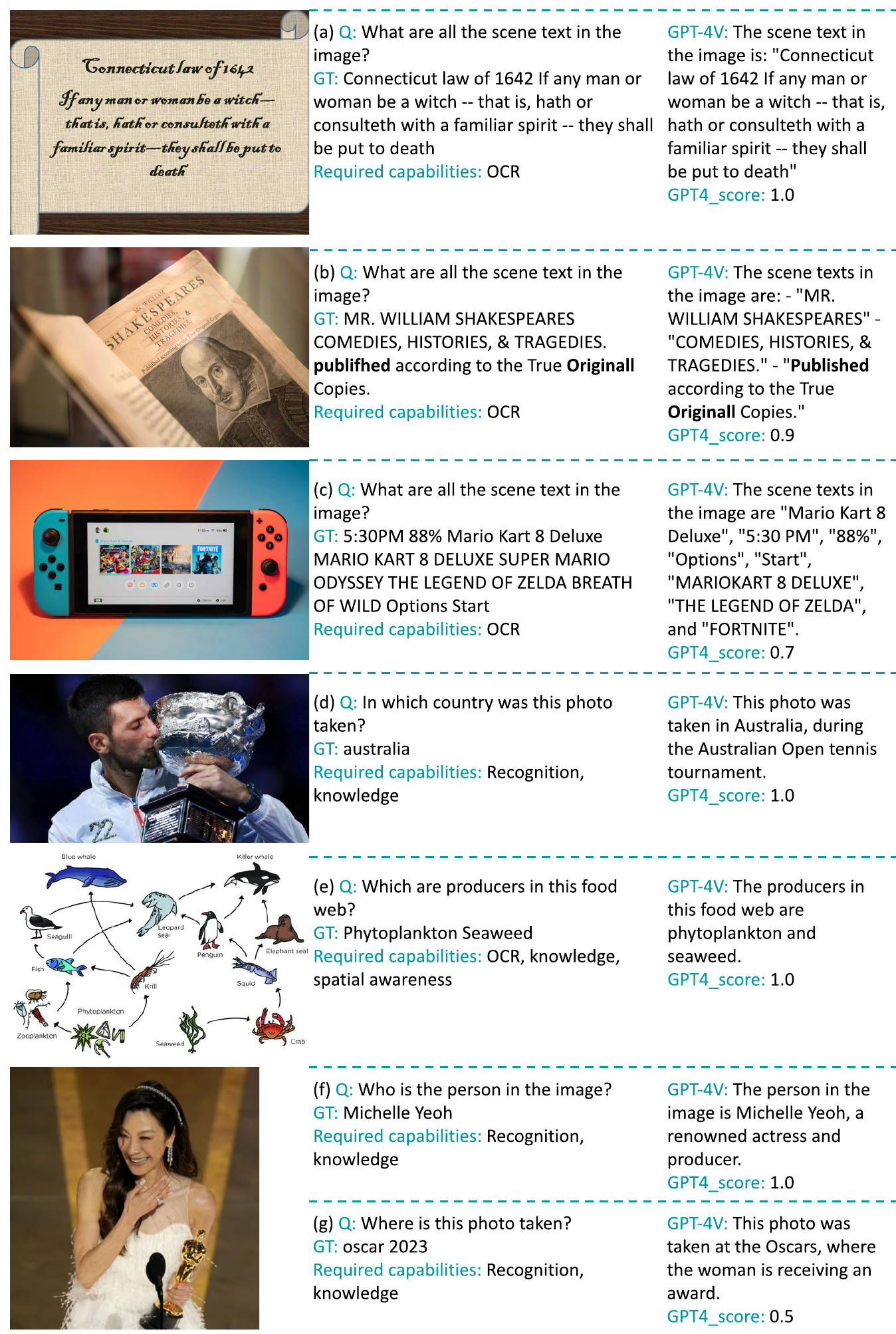}}
\caption{GPT-4V representative success and failure cases.
}
\label{fig:gpt4v_3}
\end{figure*}

\begin{figure*}[h!]
\centering
\centerline{\includegraphics[width=0.8\textwidth]{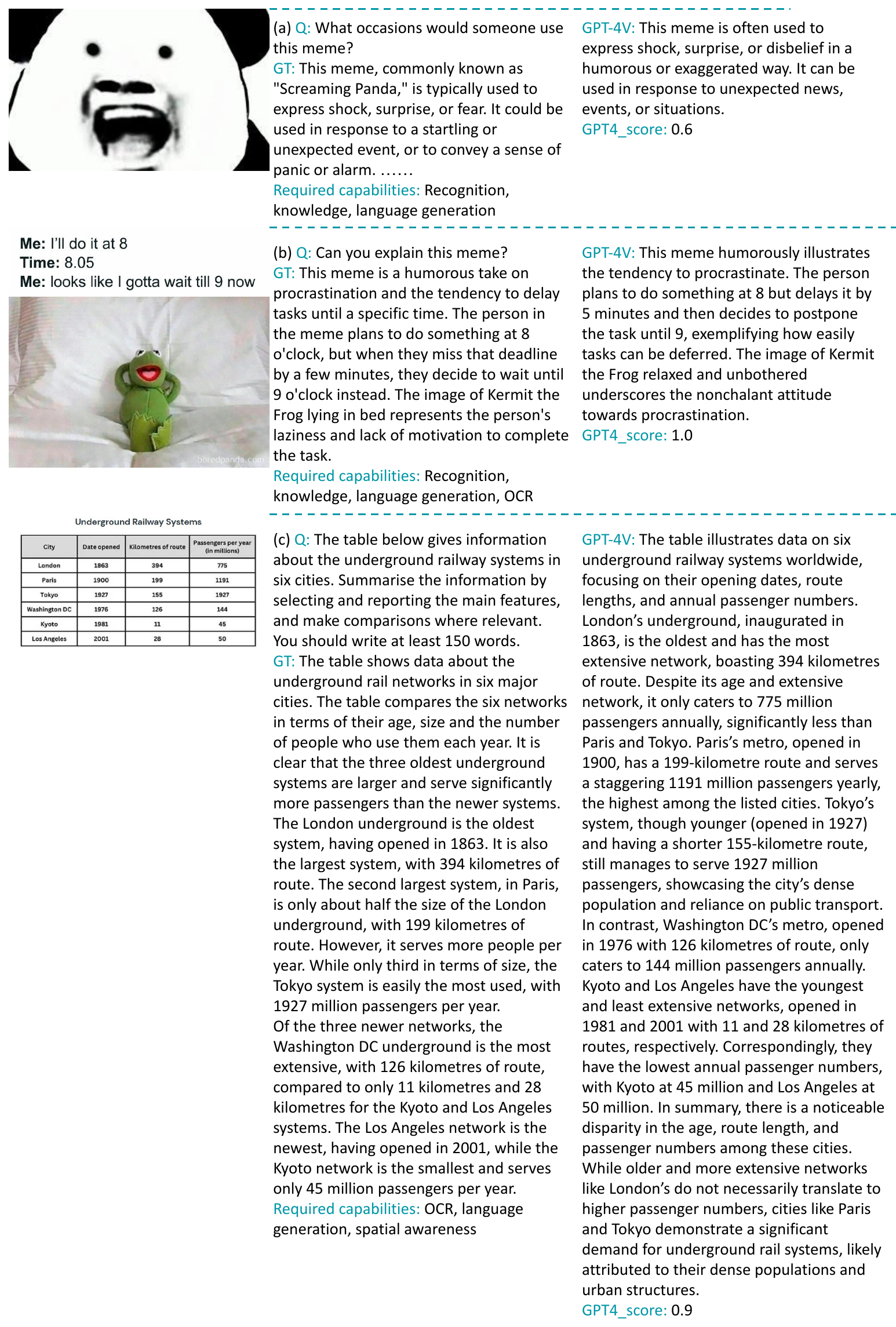}}
\caption{GPT-4V representative success and failure cases.
}
\label{fig:gpt4v_4}
\end{figure*}

\end{document}